\documentclass[twocolumn]{article}
\usepackage{graphicx}
\usepackage[left=1.5cm,right=1.5cm,top=3cm,bottom=4cm]{geometry}
\usepackage{authblk}
\usepackage{amsmath}
\usepackage{amsfonts}
\usepackage{amssymb}
\usepackage[most]{tcolorbox}
\usepackage{hyperref}
\usepackage{pifont}
\usepackage{array}
\usepackage{transparent}
\usepackage{tabularx}
\usepackage{booktabs} 

\usepackage{soul}
\usepackage{float} 
\usepackage[numbers,sort&compress]{natbib}
\usepackage[version=4]{mhchem}  

\usepackage{titlesec}
\usepackage{ifthen}

\usepackage{multirow, booktabs}

\usepackage{array}
\newcolumntype{P}[1]{>{\centering\arraybackslash}p{#1}}

\title{Reinforcement Learning for Sustainable Energy: A Survey}
\author[ ]{Koen Ponse*, Felix Kleuker*, Márton Fejér*, Álvaro Serra-Gómez, Aske Plaat, Thomas Moerland}

\date{}
\affil[ ]{Leiden University}
\affil[*]{Equal contribution}

\begin{document}

\maketitle

\begin{abstract}
The transition to {\it sustainable energy} is a key challenge of our time, requiring modifications in the entire pipeline of energy production, storage, transmission, and consumption. 
At every stage, new sequential decision-making challenges emerge, ranging from the operation of wind farms to the management of electrical grids or the scheduling of electric vehicle charging stations. 
All such problems are well suited for {\it reinforcement learning}, the branch of machine learning that learns behavior from data. 
Therefore, numerous studies have explored the use of reinforcement learning for sustainable energy. 
This paper surveys this literature with the intention of bridging both the underlying research communities: energy and machine learning. 
After a brief introduction of both fields, we systematically list relevant sustainability challenges, how they can be modeled as a reinforcement learning problem, and what solution approaches currently exist in the literature. 
Afterwards, we zoom out and identify overarching reinforcement learning themes that appear throughout sustainability, such as multi-agent, offline, and safe reinforcement learning. 
Lastly, we also cover standardization of environments, which will be crucial for connecting both research fields, and highlight potential directions for future work. 
In summary, this survey provides an extensive overview of reinforcement learning methods for sustainable energy, which may play a vital role in the energy transition.

\end{abstract}

\section{Introduction}

\begin{figure}[ht]
    \centering
    \includegraphics[width=0.5\textwidth]{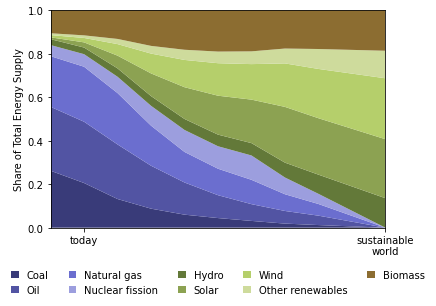}
    \caption{Potential development of the energy mix from today towards full sustainability (trajectories are not fully based on real projective data and are just for illustrative purposes) 
    }
    \label{fig:primary-energy-shift}
\end{figure}

Driven by population growth and higher per capita power use, an already rising global power demand is expected to increase further in the coming years. According to the Statistical Review of World Energy 2023 \cite{StatisticalReviewofWorldEnergy}, currently more than 70\% of primary energy\footnote{Primary energy is a measure of the energy we use, categorized under its original form found in nature.} is derived from fossil fuels.  The current share of fossil fuels in the existing energy landscape needs to be replaced by sustainable counterparts to mitigate environmental effects, such as global warming \cite{UNFCCC2018}, and to be less dependent on finite world resources. More generally, we need to \textit{meet the }(energy)\textit{ needs of the present without compromising the ability of future generations to meet their own needs} \cite{kutscher2018principles} | a common way to define sustainable energy. In other words, we must replace non-sustainable primary energy sources with suitable sustainable counterparts, as depicted in Figure \ref{fig:primary-energy-shift}. 

The backbone of scaling up the total energy supply in a sustainable landscape are renewable energy sources, such as wind or solar. However, in contrast to fossil power plants, these sources do not provide the energy on demand, which poses a variety of optimization challenges. 
While traditionally we were to fit the supply according to demand, with renewable energy sources, the supply needs to be allocated in an optimal way to guarantee the best usage and stability of electrical grids. 
Components within a grid are faced with new optimization challenges as well. For example, storage systems such as batteries could help balance the load on the grid. Another big avenue is the electrification of the transport sector, further intensifying the challenge of optimal energy distribution, e.g. in the form of EV Charging. The shift away from fossil fuels also requires a tremendous scale-up of installed capacity of renewable energy sources, rendering optimal installation and operation of these even more important. 

Reinforcement learning, a major branch of Machine Learning, aims to solve sequential decision tasks and is well suited to address many of the optimization and control challenges in the field of sustainable energy. 
Machine learning in general would help in many regards in mitigating test time costs and recognizing patterns in large amounts of data. 
Reinforcement learning in particular can aid in finding optimal strategies, usually called policies, just by interacting with an environment and receiving partial feedback.
In contrast to supervised learning, reinforcement learning does not require control actions to be individually labeled, but instead learns from the outcomes of its actions through trail-and-error. This in turn allows it to potentially outperform human solutions.
As such, reinforcement learning is used in the field of sustainable energy with the hope of significantly enhancing the efficiency and reliability of these systems \cite{yang2020reinforcement}.

This survey aims to connect the machine learning (reinforcement learning) community with the sustainable energy community. As such, we will be approaching the field from both angles in different sections. For researchers coming from the machine learning domain, we group challenges in the sustainable energy field in an easy-to-understand taxonomy and point to research that has been undertaken to address these challenges. Researchers from the energy domain are presented with a general introduction into reinforcement learning, pointing to possibly relevant literature. Additionally, we group the currently existing literature in reinforcement learning challenges that need to be addressed in the various energy problems.

We find that the field is still relatively young, and major reinforcement learning literature has largely not yet found its way into applied research. Notable topics that would require more work, in order to get to real-world deployment, are safe reinforcement learning and offline reinforcement learning.
Furthermore, we observe a very wide and diverse number of benchmarks (environments) in the research field, while we should strive to more standardization. This standardization may prove to be the most important bridge for the two research fields to come together. 
When this happens, we see a lot of potential for reinforcement learning in the energy field.

Note that the transition to sustainable energy involves both technical and economic challenges. For this survey, we focus on the former, discussing the technical aspects of generation, storage, transport, and consumption of sustainable energy. Of course, practical deployment will also require economic tools, such as sustainable energy trading, logistics, and scheduling. However, these challenges are not specific to the energy transition and therefore fall outside the scope of this survey.

Our contributions are as follows: (1) We provide an overview of the whole energy chain and possible options for reinforcement learning optimization within this chain. We aim to do this in a way that is (2) specifically designed to connect the two research fields, the energy field and the machine-learning field, by repeatedly switching focus points. Lastly (3), we identify pitfalls, bottlenecks, and promising directions for future research.

\vspace{0.2cm}
The remainder of this survey is structured as follows. We start with an overview of related surveys on machine learning and sustainable energy, and how they compare to the present paper (Section \ref{sec:related_work}). Then, Sections \ref{sec:2} and \ref{sec:3} provide a broad overview of both key topics: Section \ref{sec:2} introduces the sustainable energy landscape (intended for researchers from the reinforcement learning community), while Section \ref{sec:3} present an overview of reinforcement learning for energy researchers. Next, The core of the survey follows in Sections \ref{sec:4} and \ref{sec:5}. The former, Section \ref{sec:4}, surveys the full range of sustainable energy applications in which reinforcement learning methods have been applied. The structure here comes from the energy side, focusing on production, storage, transport and consumption of sustainable energy. Then, Section \ref{sec:5} again flips the view to the reinforcement learning perspective, discussing the overarching reinforcement learning themes we encounter along the full sustainable energy chain. Afterwards, \ref{sec:6} discusses benchmarking and performance metrics, a major topic in all of machine learning, and a crucial topic in the bridge between both fields. Lastly, Section \ref{sec:7} summarizes and discusses our findings, including recommendations and possible directions for future research.

\section{Related Work} \label{sec:related_work}

Various articles have surveyed the use of machine learning for sustainable energy \citep{rolnick2022tackling,yao2023machine, donti2021machine,rangel2021machine,ifaei2023sustainable,ahmad2022data,perera2014machine}. In general, these surveys identify much potential for machine learning methods in the sustainable energy transition, on a wide range of applications. 
However, these overviews primarily focus on {\it supervised learning} techniques, where we try to forecast certain properties, such as climate predictions. 
Although some surveys include {\it reinforcement learning} and {\it control} methods \citep{rolnick2022tackling}, this is generally not the main focus.

Several surveys do specifically cover reinforcement learning methods for sustainable energy \citep{yang2020reinforcement,cao2020reinforcement,vazquez-caRein2019}, but these typically zoom in on a specific subfield of the entire sustainability pipeline, such as the electricity grid \citep{yang2020reinforcement,cao2020reinforcement} or demand response \citep{vazquez-caRein2019}. In contrast, the present survey covers all steps of sustainability, from production (e.g., solar panels, wind farms) to storage (e.g., hydrogen), transport (e.g., electricity grids) and consumption (e.g., smart buildings, electrical vehicles). The present paper thereby provides an integrated view of the entire sustainability chain, whose individual challenges are often closely intertwined.

There are two additional motivations for our survey. First of all, the developments in machine learning for sustainable energy move incredibly fast. A search using the Arxiv API with keywords "Reinforcement Learning" and "Sustainable Energy" reveals 1798 papers, of which 1486 (83\%) have been published in or after 2020.
The field has therefore moved incredibly fast, and previous surveys of reinforcement learning and sustainable energy \citep{yang2020reinforcement,cao2020reinforcement,vazquez-caRein2019} have not covered this large part of the literature. 

Moreover, we also observe that previous overview papers predominantly originate from the `energy literature' \citep{cao2020reinforcement,vazquez-caRein2019,ifaei2023sustainable,ahmad2022data}. Machine learning and sustainable energy are of course two separate research fields, and the bridge between two communities often takes effort from both sides. In general, we observe energy researchers have started exploring machine learning techniques for their problems, but pure machine learning researchers have more trouble entering the field, probably because they lack clear benchmarks and problem definitions (see Section \ref{sec:6} as well). An additional goal of this survey is therefore to provide a bridge from the machine learning perspective, also terminology-wise | in hope of finding common ground.

\section{Areas of Sustainable Energy}
\label{sec:2}

\begin{figure*}[ht]
    \centering
    \includegraphics[width=0.95\textwidth]{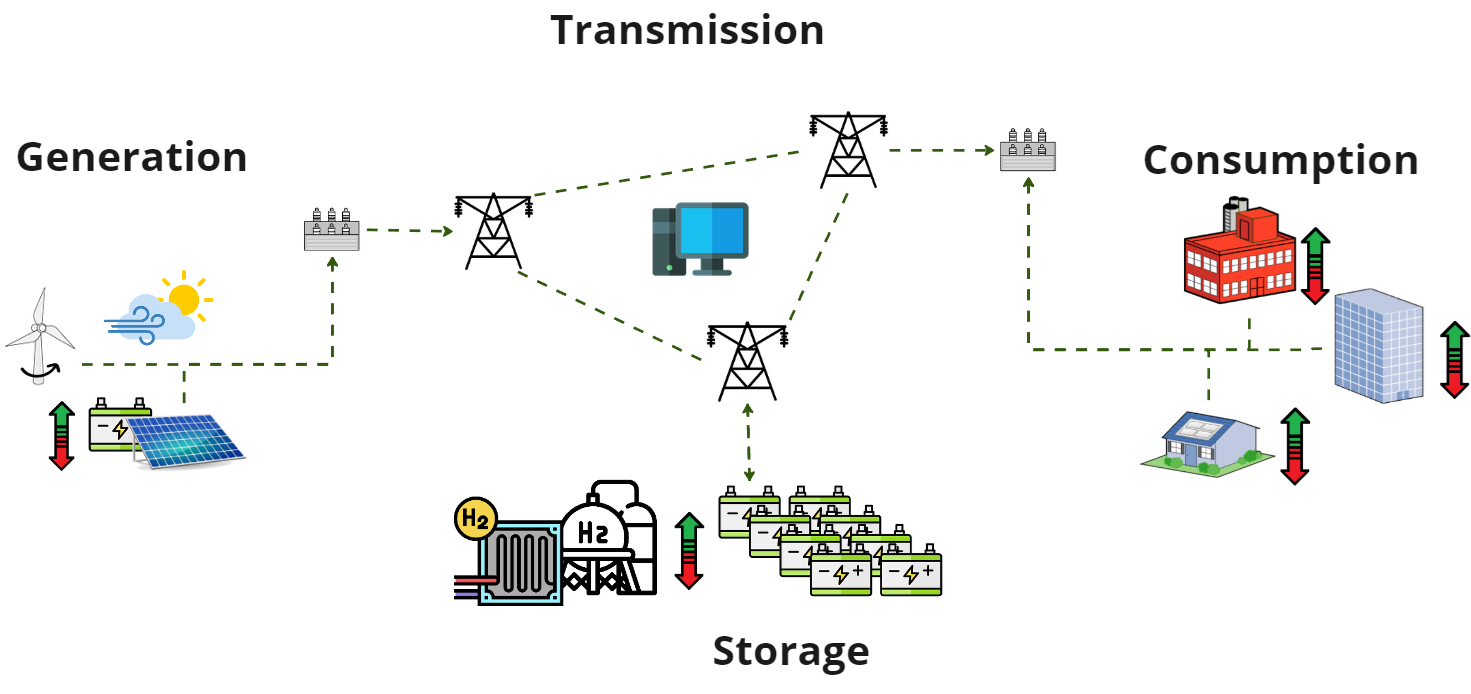}
    \caption{Overview of energy systems and our taxonomy used throughout this survey. We separate sustainable energy systems in one of four pillars: generation, storage, consumption and transmission. Notably, transmission links all the various energy areas. This taxonomy will be used primarily in Sections \ref{sec:3} and \ref{sec:4}, where we briefly introduce each pillars, and dive deeper into the active research in each of the pillars, respectively.}
    \label{fig:taxonomy_overview}
\end{figure*}

Energy-related processes often follow the same pattern: energy is produced, stored, transported, and finally consumed.
This segmentation of energy provides a natural taxonomy, as shown in Figure \ref{fig:taxonomy_overview}. All four areas of this taxonomy provide distinct challenges and opportunities for sustainability improvements through reinforcement learning. 
The remainder of this section introduces the main set of challenges that appear in each of the four areas.
Note that although we attempt to separate problems into their respective areas for clarity, real-world problems may deal with multiple parts of the energy chain and need to be optimized together.

\subsection*{Generation}

To ensure that our society is powered by sustainable energy, it is important that the energy produced comes from renewable sources. These sources include (1) Hydropower, which produces electricity primarily by converting the potential energy of water, stored in reservoirs, through dam infrastructure.
(2) Solar power, producing electricity from solar radiation. (3) Wind power, capturing kinetic energy to convert into electricity. More niche areas include (4) Tidal and (5) Geothermal power.
Furthermore, because of its closed carbon cycle, (6) energy from biomass is also considered renewable. Lastly, (7) nuclear fusion would feature enough characteristics of renewable energy, due to the plentiful supply of fuel (hydrogen) and the absence of harmful long-term waste products | an attribute that distinguishes it from nuclear fission, which we exclude from our study due to these inherent byproducts. Similarly, we exclude any endeavors solely aimed at optimizing fossil power plants from this study.

\subsection*{Storage}

Energy is often produced non-locally, requiring us to transport it in time (storage) and space (transmission) to reach the consumer at the right time and location.
Storing electricity requires us to convert it into a different form of potential energy. For example by pumping water into lakes of hydro power facilities. Electricity can also be stored in chemical (inner) energy, like batteries or hydrogen. Often, geographical factors such as natural height and access to water heavily influence the choice of energy storage.
This survey includes EV batteries and high capacity storage, but excludes small-scale batteries because the immediate impact on sustainable energy is not known.

\subsection*{Transmission}
Sustainable energy is primarily transported over {\it electricity grids}, which move electrical energy from producers to consumers over networks of transmission lines \cite{escobar2021}. 
The key challenge of energy grids is to match {\it supply} and {\it demand}, while at the same time ensuring grid stability (e.g., {\it voltage} and {\it frequency control} to ensure both stay within safe bounds). Demand is relatively fixed,  although we may try to influence it through {\it demand response}, for example by varying the price of electricity throughout the day. However, the main challenge of grids is to match supply to the demand, a challenge that gets aggravated with sustainable energy sources. For example, a traditional fossil power plant can produce electricity at any required moment, but solar power is only available when the sun is actually shining. This generates challenges for energy {\it dispatch} (when and where do we release energy into the grid) and {\it energy management} (how do we locally store and route energy to efficiently operate a (mini-)grid). All these challenges get pronounced with sustainable energy sources, since they are inherently distributed (spread out) and have variable and uncertain production profiles \cite{escobar2021, Li2023, hasan2023, dileep2020, Salam2024, jirdehi2020}. 

Finally, note that other means of energy storage, such as hydrogen, also require other means of transmission, for example by trucks or pipelines. 
However, due to the variability of sustainable energy production, leading to a more significant \textit{duck curve} \citep{sheha2020solving}, electricity grids require a significant change in control operation. 
This change is not something we naturally see in trucks or pipes. In turn we focus on electricity grids as the primary technical challenge in transmission of sustainable energy. 

\subsection*{Consumption}

Historically, we have mostly balanced energy supply and demand through supply (aided by storage solutions). However, the rise in global electricity useage and ongoing electrification of society drive us to seek for more effective methods of managing and reducing energy demand. With recent increases in energy prices, such optimizations are likely to become a growing field of interest.

Typical consumers of energy include \textit{houses and offices}, where the cost of energy is primarily driven by heating and cooling installations.
Different challenges are found in what we categorize as \textit{mobile consumers}, encompassing mainly electric vehicles (EV's) and their required charging stations.
Finally, our last category of energy consumers is the \textit{industry} sector. Here, reinforcement learning sustainable energy methods may be designed to address highly specific needs, which is justified by the volume of energy use.

\section{Reinforcement Learning Basics} \label{sec:3}

\begin{table*}[ht] 
    \centering
    \caption{Example of potential (simple) MDP definitions for two sustainable energy tasks.}
    \vspace{0.15cm}
    \label{tab:mdp-examples}
    \begin{tabular}
    {p{0.10\textwidth}|p{0.25\textwidth}|p{0.25\textwidth}|p{0.25\textwidth}} 
         & State space ($\mathcal{S}$) & Action space ($\mathcal{A}$) & Reward ($r$) \\ \hline
        Wind & Current angle of the rotor of the turbine $\theta$, wind speed $v$ and direction $\phi$ & change in the rotor angle $\Delta \theta$ & energy produced $E(\theta,v,\phi)$ \\ \hline 
        Building Control & Occupancy, weather data, energy price & Heating temperature setpoint & Minimize energy cost, while maintaining temperature threshold \\ 
    \end{tabular}
\end{table*}

In many of the domains that are discussed in the previous section, we are faced with optimization problems such as maximizing energy generation, minimizing power usage, or optimizing power allocation.
These optimization tasks often require us to decide on multiple actions to obtain good average performance over a long time horizon.
These problems are known as {\it sequential decision-making problems}, for which we may employ reinforcement learning.
Reinforcement learning is a machine learning approach for finding an optimal policy by interacting with an environment. It is often used for sequential decision-making problems, where actions from the past influence states into the future.

Informally, reinforcement learning problems consist of an agent and an environment. The environment is in a state, and after an agent chooses an action, the environment follows a transition function to determine both the new state and a reward, a numerical value indicating how "good" the new state is. The goal of the agent is to find a so-called policy of optimal actions for each state, thereby solving the reinforcement learning problem by sampling the environment with its actions. 

More formally, the sequential decision-making problem can mathematically be defined as a Markov decision process (MDP) \cite{littman1994markov,suttonbartobook2018} 
, defined as a tuple ($\mathcal{S},\mathcal{A},p,r,p_0,\gamma$). The state space $\mathcal{S}$ is the set of all states, the action space $\mathcal{A}$ is the set of all possible actions, $p$ is the transition dynamics distribution 
$p: \mathcal{S} \times \mathcal{A} \to \Delta(\mathcal{S}), (s,a) \mapsto p(s'|s,a)$, $r$ is the reward function that maps transitions into rewards $r: \mathcal{S} \times \mathcal{A} \times \mathcal{S} \to \mathbb{R}$, $p_0 \in \Delta(\mathcal{S})$ is the initial state distribution and $\gamma \in (0,1)$ is the discount factor, which governs the importance of future rewards. 
In an MDP, the transition dynamics distribution is Markovian, meaning that the transition to a next state, given an action only depends on the current state, i.e. $p(s_{t+1}|s_t,a_t,\dots,a_0,s_0) = p(s_{t+1}|s_t,a_t)$. To provide some intuition, we briefly present some examples on how to formulate energy problems as an MDP in Table \ref{tab:mdp-examples}.

To select actions in any given state, a policy is used, that is, a mapping $\pi: \mathcal{S} \to \Delta(\mathcal{A}), s \mapsto \pi(a|s)$. Alternating sequences of states and actions are usually denoted as trajectories $\tau = (s_0,a_0,s_1,\dots,s_T)$, and each policy induces a distribution $p_\pi(\tau)$ over such trajectories in an MDP.
For a given trajectory, the return $G_t$ is defined as the total discounted rewards from time-step $t$ (resp. state $s = s_t$) onwards
\begin{equation*}
    G_{t} = \sum_{k=t}^\infty \gamma^{k-t} \, r(s_k,a_k,s_{k+1}).
\end{equation*}
Value functions for a given state or a state-action pair are defined as the expected return given a fixed state or a state-action pair, respectively:
\begin{align*}
    V_\pi(s) &= \mathbb{E}_{\pi} [G_t | s_t = s] \\
    Q_\pi(s, a) &= \mathbb{E}_{\pi} [G_t | s_t = s, a_t = a].
\end{align*}
Solving an MDP is defined as finding a policy $\pi^\ast$ such that it maximizes the expected return of trajectories:
\begin{equation*}
    \pi^\ast \in \arg\max_\pi \mathbb{E}_{\tau \sim p_\pi} [V_\pi(s_0)].
\end{equation*}

Finding an optimal policy is usually done in an iterative fashion: Knowledge about the values $Q_\pi$ under a given policy can be used to improve the policy, for example, by acting greedily w.r.t. the current values, naturally being denoted as a \textit{policy improvement} step. This, in turn, requires a new evaluation of the values under this new policy, a step usually called \textit{policy evaluation}. Generally, this scheme of alternating between policy evaluation and policy improvement is called Generalised Policy Iteration (GPI).

The policy evaluation step can be done by utilizing the so-called Bellman-equations \cite{bellmanMark1957}, looping over all state-action pairs and updating the value functions; a method known as Dynamic Programming \cite{suttonbartobook2018}. However, solving an MDP with Dynamic Programming requires access to a known model of the environment, That is, $p$ and $r$ must be known, which, for most real-world applications, is not the case. 

In reinforcement learning, we therefore assume a setting in which the transition dynamics $p$ and reward function $r$ are unknown. 
Additionally, instead of looping over each state-action pair, in reinforcement learning, we allow an agent to collect experiences, i.e. trajectories, by interacting with an environment following a policy $\pi$.
These experiences can then be used to learn value functions $V_\pi$ or $Q_\pi$, in what is known as \textit{value-based} methods, or directly learn an explicit policy $\pi$, in what is known as \textit{policy-based} methods.
Some methods use the obtained experiences to learn both the values and an explicit policy. These methods belong to the class of \textit{actor-critic} \cite{actor-critic_paper} methods and are particularly popular in some of the current state-of-the-art algorithms \cite{ppo_paper, td3_paper, ddpg_paper, trpo_paper, SAC_paper, muesli_paper}.

Well-known value-based methods, such as Q-learning \cite{qlearning_paper} and SARSA \cite{sarsa_paper}, store their learned value functions in a table, in computer memory. As a table entry is required for each state-action pair, these tabular reinforcement learning methods may quickly become too memory intensive for realistic applications.
To combat this, we use function approximations that generalize over the state-action space, building up on the advancements made in Deep Learning. The value functions \cite{dqn_paper, rainbow_paper} or policies \cite{reinforce_paper} are encoded by a parameterized function, generally neural networks. This field is now known as \textit{deep reinforcement learning}, and has seen numerous successes over the years \cite{silverMast2016, silverMast2017a, kaufmannCham2023a, christianoDeep2023, vinyalsGran2019, openaiDota2019}.

Various sub-fields in reinforcement learning have emerged for solving problems encountered in different environments.
Most notably, since MDPs require the transition dynamics to be Markovian, the states are required to contain all relevant information. However, in many problems, not all information is known (hidden or stochastic information).
Such environments are said to be partially observable MDPs (POMDPs) \cite{moradPOPG2023, pleinesMemo2022} and other methods exist to approximately solve them with reinforcement learning.
Furthermore, while (good) simulators may not always be available, datasets of actions, observations, and rewards may have been generated by real-world installations. Applying reinforcement learning on this dataset, without a simulator, is known as offline reinforcement learning \cite{levineOffl2020, prudencioSurv2024}. 

In particular, {\em model-based} algorithms \cite{moerlandMode2023,plaat2023high} may work well in offline reinforcement learning problems. Unlike model-free algorithms, model-based algorithms do have access to the transition dynamics and rewards, either via a known or a learned model.

Reinforcement learning is a rich and active field of research, and there are more methods for different problems. Although we provide some more explanation on some subfields in Section \ref{sec:5}, explicit explanation of every type of problem is beyond the scope of this survey. However, various other surveys exist on topics that come up in later sections, such as multi-agent reinforcement learning \cite{wongDeep2023a}, exploration \cite{ladoszExpl2022}, and safe RL \cite{guRevi2023}. 
For a comprehensive introduction to tabular reinforcement learning, we recommend Sutton \& Barto \cite{suttonbartobook2018} and Kaelbling et al. \cite{kaelbling1996reinforcement}, for deep reinforcement learning we recommend Francois-Lavet et al. \cite{francois-lIntr2018} and Plaat \cite{plaat2022deep}.

\section{Applications of Reinforcement Learning in Sustainable Energy} \label{sec:4}

This section focuses on the use of reinforcement learning, as introduced in Section \ref{sec:3}, in the different sustainable energy areas that are introduced in Section \ref{sec:2}.
We will go over each area of our taxonomy, discuss sub-areas, and highlight problems that are currently being worked on together with how researchers have so far addressed these problems.

First, Section \ref{sect:generation} covers generation: predominantly hydro, solar and wind, and also smaller fields such as tidal, biomass, fusion and geothermal.
Section \ref{sect:storage} dives into reinforcement learning approaches in energy storage solutions, discussing batteries, hydrogen and pumped hydro storage.
Next, Section \ref{sect:consumption} discusses ways optimize energy consumption and help balance the grid from the energy demand side. Buildings, electric vehicles and the industry sector will be featured here.
Finally, Section \ref{sect:grids} discusses energy grids, which play a crucial role in connecting all previous components (Figure \ref{fig:taxonomy_overview}). Although all components could be optimized/learned together, grid literature typically assumes the other components have some static controller.

\subsection{Generation} \label{sect:generation}

A natural angle to approach sustainable energy is to improve the efficiency of inherently sustainable sources, mentioned in Section \ref{sec:2}. Reinforcement learning can help in optimizing the control and operation of such energy generation facilities, thereby boosting the efficiency. This, in turn, would allow to increase the share of these sources and consequently shift the primary energy landscape in the desired direction.

\begin{figure}[H]
    \centering
    \includegraphics[width=0.45\textwidth]{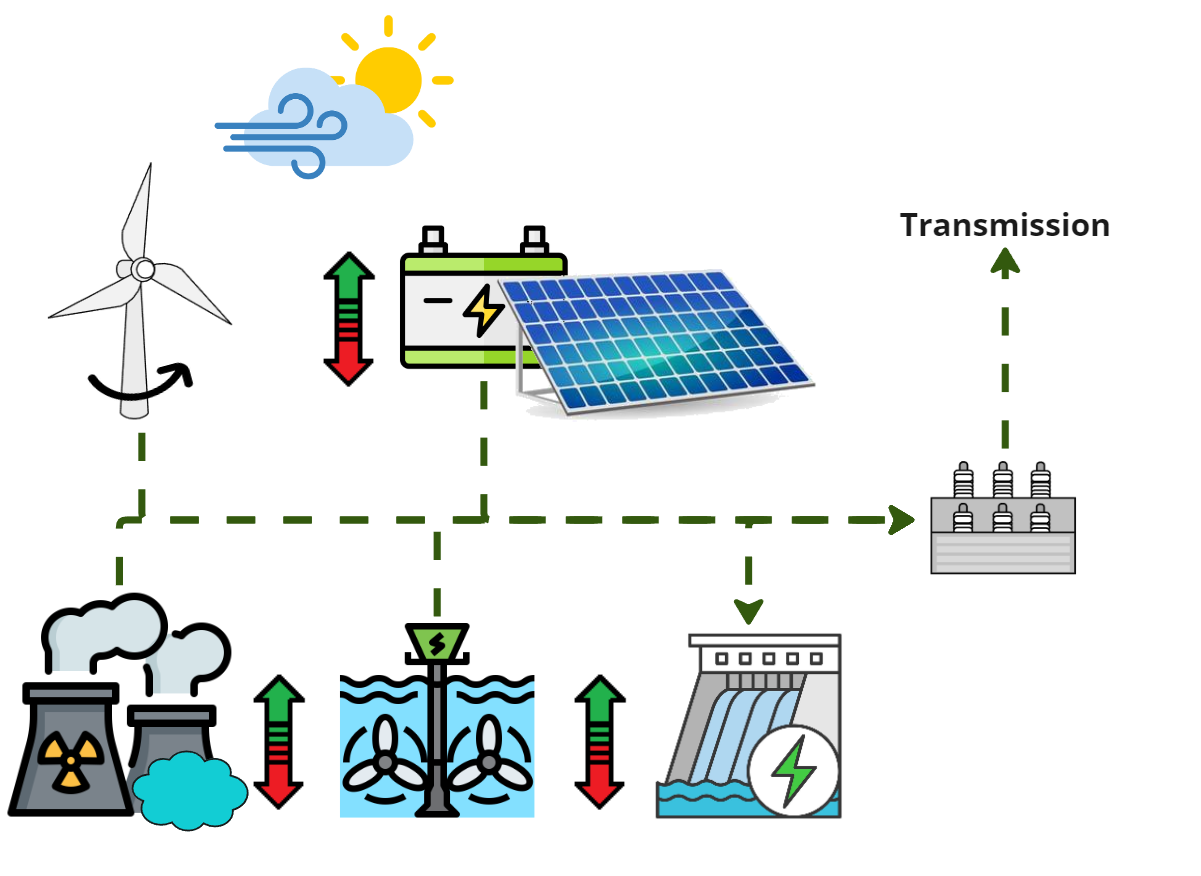}
    \caption{Overview of sustainable power generation, the first pillar in our taxonomy.}
    \label{fig:gen}
\end{figure}

\subsubsection*{Hydropower}

Hydropower accounts for 16\% of the globally produced electricity (not to be confused with energy), contributing the largest share of all renewable energy sources \cite{IHA_2022, StatisticalReviewofWorldEnergy}. Although hydropower is an excellent source of sustainable energy, geological requirements make it comparatively less scalable compared to other renewable sources, such as wind and solar energy. Economically viable hydropower potential in some areas has now largely been exploited \cite{IHA_2022, kleiven_revisiting_nodate}, and the further enhancement would require substantial investment. 
Kleiven et al. \cite{kleiven_revisiting_nodate} propose an investment model using reinforcement learning to determine an optimal upgrade capacity along with its optimal point in time; this decision is based on projected electricity prices and water inflows, modeled as a Markov Decision Process.

Other models aim to improve the economic viability and efficiency of hydropower plants. Xu et al. \cite{xu_deep_2021} propose a deep Q network method, where the water level and inflow represent the state space. The objective is to maximize the total generated energy by adjusting the release of water.
Another work has investigated maximizing total generated electricity in a multi-reservoir (yet single-agent) setting \cite{mitjana2022managing}.
A similar approach, focused on optimizing total revenue earned is also proposed \cite{riemer-sorensen_deep_2020}. Here, the generated energy is multiplied by a forecasted electricity price, introducing an additional element of uncertainty.

\subsubsection*{Wind Power}

With a 7\% share of global electricity production, wind power accounts for the second highest production of renewable sources \cite{StatisticalReviewofWorldEnergy}. Due to variability in wind speed (and angle), a primary challenge in the optimization of wind power plants lies in accurately forecasting wind speed (shifts in direction can usually be adjusted for in real time). 
Recently, reinforcement learning has been proposed for these prediction tasks \cite{liu_new_2020, jalali_new_2022}, as apposed to supervised learning that has been the standard.
When coupled with a battery system to compensate for periods of low wind, reinforcement learning can find policies to maintain a stable power supply based on these wind predictions \cite{yang_reinforcement_2021}.

Note that it is important not only to adapt to future wind speed predictions, but also to optimize power output and battery charging load in real-time, responding to current wind conditions. This is sometimes referred to as Maximum Power-Point Tracking, presenting the optimal load to the generator depending on wind conditions \cite{wei_adaptive_2016}.  Reinforcement learning has demonstrated promising results in finding optimal policies under variable wind conditions \cite{wei_adaptive_2016, fernandez-gauna_actor-critic_2022}. 
Furthermore, the application of reinforcement learning allows the scaling up of optimization parameters without significantly increasing the inference time, enabling innovative design proposals for wind turbines \cite{jia_reinforcement_2021}.

Lastly, note that, as wind farms grow in size, the complexity of the optimization challenge also intensifies. The optimal operational point for this group of turbines is usually not a linear aggregate of the optima of the individual turbine due to interaction effects, such as one turbine being in the wind shadow of another turbine. To identify the optimal operational point for the entire collection under specific wind conditions, multi-agent reinforcement learning methods may offer more optimal solutions \cite{bui_distributed_2020}.

\subsubsection*{Solar Power}

Electricity generated from solar power constitutes approximately 4.5\% of global electricity production, positioning it as the third most prolific renewable source \cite{StatisticalReviewofWorldEnergy, IHA_2022}. However, solar power might be the best candidate for scaling up because of the abundance of untapped potential to deploy photovoltaic (PV) cells.

A widely discussed subject in the realm of reinforcement learning for solar power is again the so-called Maximum Power-Point Tracking, where the aim is to maximize the power produced of a set of PV cells under non-optimal conditions, such as shading \cite{ding_global_2019, lin_self-tuning_2021, chou_maximum_2019, phan_deep_2020, zhang_memetic_2019, kofinas_reinforcement_2017, singh_reinforcement_2021, yadav_deep_2022}.
Photovoltaic systems have a unique global optimum under ideal conditions, and multiple local optima otherwise. The challenge lies in adjusting the controllable parameters, such as the voltage, to maintain the operational point at the global maximum. All models incorporate electric current and voltage in their state spaces, yet some studies incorporate additional variables, such as solar irradiance and temperature \cite{chou_maximum_2019}. Interestingly, while early work in this field employed a tabular reinforcement learning approach \cite{ding_global_2019, lin_self-tuning_2021, chou_maximum_2019, phan_deep_2020, zhang_memetic_2019, kofinas_reinforcement_2017}, more recent work mainly adopts deep reinforcement learning techniques \cite{singh_reinforcement_2021, yadav_deep_2022}. 

Although these algorithms target efficiency improvements in isolated photovoltaic systems, there are also approaches that optimize the performance of photovoltaic systems equipped with batteries and grid access. Optimization, in this context, may involve maintaining a specified battery level \cite{shresthamali_adaptive_2017} or achieving energy neutrality in an energy network \cite{ge_maximizing_2021}.

Alternatively to photovoltaic systems, which convert solar radiation into electricity, there are systems that harvest the heat energy of solar radiation. For example, we may use reinforcement learning to optimally control a heliostat field to convert sunlight into heat (and subsequently into power) \cite{zeng_real-time_2022}. Another study investigates solar fields that generate hot water \cite{correa-jullian_operation_2020}. Lastly, in addition to maximizing power output, reduction of maintenance cost can be crucial to increase the economic efficiency. In this context, one study uses a reinforcement learning approach for fault detection and diagnosis, extracting an optimal strategy through tabular Q-learning \cite{zhang_reinforcement_2020}.

\subsubsection*{Tidal Power}
In terms of installed capacity, tidal energy is less significant compared to solar, wind and hydro. However, we have seen a significant amount of reinforcement learning research in this technology.
Three primary technologies are used to generate energy from tidal flows. (1) \textit{Tidal turbines} are similar to their wind-driven counterparts and capture the kinetic energy inherent in tidal flows \cite{fangSolv2023}. (2) \textit{Wave energy converters} leverage wave motions, resisted by a power take-off, to convert kinetic energy into electricity \cite{anderliniCont2016}. A more specialized approach involves (3) \textit{Tidal Range Structures}, which generate power by artificially inducing a difference in water level between the ocean and a confined area. Turbines then generate energy by allowing water to balance out this discrepancy \cite{moreiraPred2022}.
As in hydropower, the installation of tidal power is limited by geographical constraints (coastal areas).

Wave energy converters consist of a small floating body subjected to wave forces, with its movements countered by an electric or hydraulic power take-off system. The controllable damping coefficient that influences the resistance of the power take-off has different optimal values, depending on the sea state \cite{anderliniCont2016}. To optimize this coefficient, Anderlini et al. \cite{anderliniCont2016} employ a tabular Q-learning for offshore wave energy converters, specifically for (heaving) point absorbers \cite{falcao2010wave}. This work was later extended for onshore wave energy converters \cite{bruzzoneRein2020}.
In more recent work, Anderlini et al. \cite{anderliniReal2020} use soft actor critic, a deep reinforcement learning algorithm to solve similar problems. Notably, newer approaches deal with more modern wave energy converters featuring multiple power take-offs, rather than one. In this more complex space, multi-agent systems have also been tried, based on PPO \cite{sarkarMult2022}. 

To a lesser extent, reinforcement learning has also found its way to tidal turbines and tidal range structures. 
PPO has been proposed for the latter to maximize the energy generation of a tidal range structure with multiple turbines, in which the inflow rate had to be controlled \cite{moreiraPred2022}.
For tidal turbines, reinforcement learning has been employed for Maximum Power-Point Tracking. A challenge analogous to problems in wind turbines, yet under different conditions.

\vspace{0.2cm}
Lastly, the three remaining energy sources | biomass, geothermal energy, and nuclear fusion | identified as inherently sustainable, are not extensively explored in the reinforcement learning literature. However, we believe that reinforcement learning has significant potential to impact control applications within these domains.

\textbf{Biomass} processes transform  biological matter from plants and animals into carbon-based energy carriers, such as ethanol or biogas (methane). Biomass processes are also denoted as Waste-to-Energy (WtE). Different conversion methods have been studied, such as thermochemical, physicochemical, and biochemical processes \cite{biomass_survey1, biomass_survey2}. The potential application of reinforcement learning in controlling these processes is presented by \textit{Faridi et al.} \cite{faridiAdva2024}. In this work, the authors introduce a model-based deep reinforcement learning approach for finding an optimal controller for a thermochemical gasification process of biomass. 
Another study \cite{lim2022reinforcement} focuses on improving the efficiency of a recovery boiler, a device designed to convert a byproduct of the paper industry into synthetic fuel. The authors used a tabular reinforcement learning approach with the specific objective of reducing the heat transfer rate.

\textbf{Geothermal Energy} systems use internal heat of the Earth to generate energy, commonly by using naturally existing high-pressure water or steam reservoirs.
Although geothermal energy systems are only available in areas with suitable seismic activity, there are various fields around the world \cite{ellisGeot1975}. 
Because of its advantage of offering a more reliable production in comparison to solar and wind power, geothermal energy offers the potential to contribute significantly to the future's sustainable energy output.
Some frameworks modeling the mechanics of a geothermal plant have been created in recent years \cite{busterNew2021, duplyakinMode2022} and these may be adapted for reinforcement learning training in the future. 
Siratovich et al. \cite{siratovichGOOM2022} have conducted small-scale experiments in one of the frameworks \cite{busterNew2021}, demonstrating that by manipulating several pressure valves, an agent managed to increase total energy output over a two and a half year period.

\textbf{Nuclear fusion} is a promising clean energy source. 
Much of the research effort of harnessing nuclear fusion is aimed at controlling the very high temperature of the fusion plasma, often in tokamaks, a type of experimental fusion reactor. Researchers are exploring automated control methods, predicting and mitigating disturbances in magnetic fields, ensuring stable and efficient plasma operations under high-pressure conditions \cite{Degrave2022, Seo2024, Seo2021, Wakatsuki2021, Wakatsuki2023, Tracey2024, Char2023}. Deep reinforcement learning techniques optimize various parameters and control schemes in tokamak plasmas, intended to improve the shape, duration, and temperature of plasma conditions \cite{Wakatsuki2021, Seo2021, Degrave2022}.

\subsection{Storage} \label{sect:storage}
While fossil fuels naturally and efficiently store energy in chemical form, allowing flexible matching of supply with demand, renewable energy sources do not. Consequently, we must improve solutions to store the energy generated from renewable sources in order to match supply and demand. 

\subsubsection*{Pumped Hydro Energy Storage}

An additional characteristic of some hydropower plants is the ability to operate as energy storage repositories. These are commonly called pumped hydro energy storage (PHES) systems and work by letting water flow down to generate electricity; and pumping water back up into a reservoir to store potential energy. The installed capacity of PHES is approximately 10\% of the aggregate installed capacity dedicated to hydropower \cite{IHA_2022}. Toufani et al. present a Markov Decision Process formulation with a focus on maximizing cashflows in converting hydropower facilities into PHES systems \cite{toufani_short-term_2022a, toufani_operational_2022b}.
A related MDP formulation forms the backbone of the research by Tubeuf et al. \cite{tubeuf_increasing_2023}, wherein they study a PHES system with specific attention to safety considerations. 
The authors develop a digital twin of the physical turbine employed for pre-training the reinforcement learning model. 

In a parallel study \cite{enyekwe_speed_2023}, the focus is on replacing a conventional PID controller (Proportional, Integral, and Derivative) for speed tracking issues with reinforcement learning, to maintain the angular velocity of the turbine as consistently as possible. A reinforcement learning-based controller necessitates fewer adjustments compared to its PID counterpart, as it autonomously adapts to diverse scenarios through the learning procedure.

\subsubsection*{Batteries} 

Batteries are perhaps the most widely recognized chemical storage system. In this work, we exclude solid-state chemical storage systems such as lithium-ion batteries on the basis of the raw materials used in their production. We refer interested readers to a survey on reinforcement learning in battery storage systems \cite{subramanyaExpl2022}.
In contrast to solid-state chemical storage systems, liquid-based chemical storage systems, specifically redox-flow batteries, are considered to be well suited to address large-scale energy storage challenges \cite{rfb1_3}, in a sustainable way.
Within this domain, Sowndarya et al. \cite{s.v.Mult2022} introduce a reinforcement learning framework, based on AlphaZero \cite{silverMast2017b}, to identify new stable candidates.
Reinforcement learning has also shown promise in the construction of models to predict redox-flow battery state variables, crucial for the widespread deployment of such systems~\cite{benahmedOpti2024}.

\subsubsection*{Hydrogen} 
Additionally, energy can also be stored in the form of gaseous chemicals, most notably hydrogen.
Unlike in the aforementioned systems, where conversion, storage, and reconversion occur within the same system, hydrogen storage processes are typically distributed. Electrolysers, responsible for the electrochemical conversion of water into hydrogen and oxygen using electricity as the energy source
\begin{equation}
    \label{eq::1}
    \ce{2 H2O -> 2 H2 + O2},
\end{equation}
face a design challenge in identifying suitable catalysts. Although we are not aware of reinforcement learning work on hydrogen electrolyses, reinforcement learning has shown promise in catalysis, which may be transferable to hydrogen systems \cite{steinmannAutonomousHighthroughputComputations2022, yoonDeepReinforcementLearning2021}. 
Other studies explore employing reinforcement learning to reduce operational costs by optimizing maintenance schedules \cite{abiolaNove2023} or enhance operational efficiency in dynamic magnet field-assisted electrolyzers \cite{purnamiDoub2024}.

By the laws of chemistry, inverting reaction (\ref{eq::1}) will release (electrical) energy; a process facilitated by devices known as fuel cells. Reinforcement learning-based approaches have been applied to improve the operational efficiency of solid oxide fuel cells \cite{liData2021} and proton-exchange membrane fuel cells\cite{liCoor2021}, among various fuel cell design examples. 
Hydrogen, a key element in these conversion processes, necessitates storage. Because of its small size, diffusion in and through other materials poses a common challenge. To address this problem, reinforcement learning-based approaches are introduced to investigate hydrogen diffusion in polymer hydrogen storages \cite{sangHydr} and metal alloys \cite{tangRein2024}.

\subsection{Consumption} \label{sect:consumption}
When considering sustainable energy solutions, we commonly investigate energy production. However, improving energy {\em consumption} practices can also contribute significantly to achieving our sustainability objectives, and again, many control problems must be addressed.
In the following sections, we consider three distinct categories of energy consumers.
We first focus on {\it buildings}, where the key functions are heating, ventilation, and air conditioning systems.
Subsequently, {\it mobile} consumption is discussed, which includes electric vehicles and their charging infrastructure. 
Finally, we discuss the {\it industrial} sector, characterized by specialized energy requirements. Certain segments of the industry sector still depend on non-sustainable energy sources for parts of their product chain. Consequently, modifications in these production chains are necessary in order to rely solely on sustainable primary energy sources in the future.

\begin{figure}
    \centering
    \includegraphics[width=0.45\textwidth]{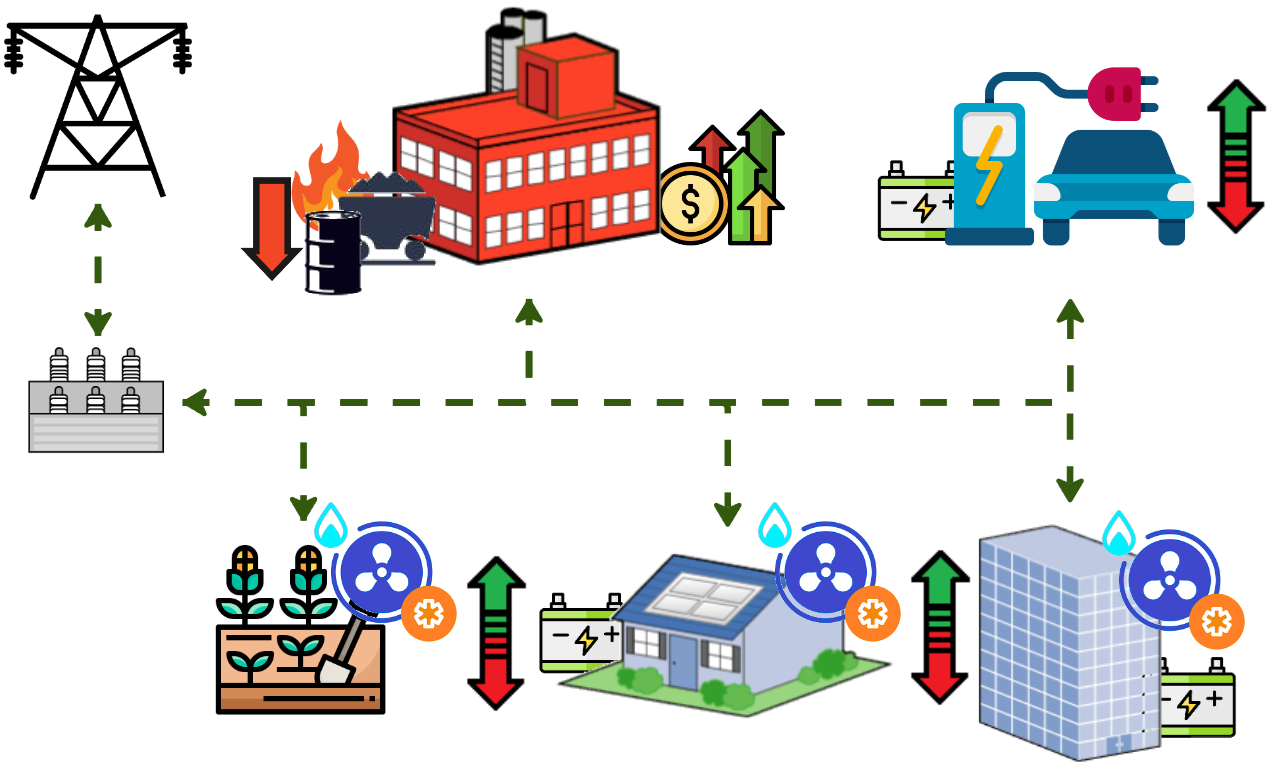}
    \caption{Overview of sustainability challenges for reinforcement learning in consumption domain. Challenges lie in sub-fields related to buildings, EV(-charging), and highly specific industrial requirements.}
    \label{fig:cons}
\end{figure}

\subsubsection*{Houses \& offices}

Globally, buildings use an estimated 30\% of final energy \cite{birolWorl2022}.
Due to this large power demand and an increasing number of smart home appliances entering people's homes, there is a growing interest in optimizing energy-intensive systems in houses \& offices. In this area, reinforcement learning has gained a lot of attention recently due to its adaptability and ability to learn without data from a reward signal.
Generally, we can subdivide the existing literature into systems that optimize a single building, and those that investigate the capability of buildings for \textit{demand response} | a topic we will also discuss in Section \ref{sect:grids}. 

A large portion of the existing literature attempts to optimize controllable elements in a single building. The goal here is usually to minimize the energy cost while maintaining a desired level of human comfort \cite{liuMult2022b, yuEner2022, fangDeep2022, esrafilianSelf2022, dengOpti2022, jiaAdva2019, luDema2019, zhangWhol2019, azuatalamRein2020, coraciOnli2021, duInte2021, anCLUE2023, gaoComp2023, jeenLow2023, zhuangData2023}. 
These works primarily focus on optimizing HVAC systems (heating, ventilation, air conditioning), as they are the largest controllable contributor to total energy consumption in buildings. 
However, other works extend the action space to a wider array of appliances \cite{yuEner2022, fuOpti2022}, while sometimes also optimizing for additional objectives such as air quality \cite{dalamagkidRein2007, valladaresEner2019, yuMult2021} or luminescence \cite{parkLigh2019}.
All works in this field vary in their use of algorithms, but almost all implement some form of model-free deep reinforcement learning, often a value-based DQN method \cite{bladData2022, dengOpti2022, esrafilianSelf2022, fangDeep2022, fuOpti2022, mocanuOnLi2019, valladaresEner2019, fuEDDQ2023, jiangBuil2021}. 
Model-based solutions are sparse \cite{anCLUE2023, jeenLow2023, gaoComp2023}, but Jeen et al. has shown potential for zero-shot model-based solutions \cite{jeenLow2023}, whereby the agent quickly adapts to any new building without the need for pre-training or a simulator, which is near-impossible to obtain for every building in the world.

Many works employ an online learning approach---based on a simulated environment, not on data---for which there exists a wide range of open source Gym-based \cite{openai_gym_original}  simulators \cite{arroyoOpen2021, dermardiroVder2023, findeisBeob2022, Henz2024, jimenez-raSine2021, mechyaiMech2024, moriyamaRein2018, scharnhorsEner2021, lukianykhiMode2019, wolfleGuid2020, zhangCOBS2020, zhangzhizzZhan2024}, some still actively developed \cite{arroyoOpen2021, Henz2024, jimenez-raSine2021}, that often leverage building simulators such as EnergyPlus \cite{crawleyEner2001}. 
However, others attempt to learn through offline data \cite{bladData2022, anCLUE2023} or partially use real data in their training loop, such as building occupancy or weather data \cite{esrafilianSelf2022, fuOpti2022, duInte2021, yuMult2021}.

Demand response systems do not take the overall capacity of the network for granted and instead investigate whether buildings can play an active role in its overall load. By timely coordination of energy usage, the idea is that buildings can collectively flatten energy peaks throughout the day. It should be noted that in demand response solutions, the energy consumption and comfort level of each individual building are still considered an important part of the optimization.
Again, different simulators exist \cite{wangAlph2021a, nweyeCity2023a}, simulating multiple buildings and their combined effect on a grid.
Of particular interest in this field is CityLearn \cite{nweyeCity2023a}, an actively developed Gym environment for demand response of a cluster of buildings. Yearly competitions are organized by the developers of CityLearn \cite{citylearn_challenge_2020, citylearn_challenge_2021, citylearn_challenge_2022}, with increased complexity and realism each year.
In CityLearn, comfort levels are assumed to always be satisfied so that building simulations can be performed in advance.
Participants in the CityLearn challenges are required to carefully manage batteries, photovoltaic panels, appliances, and HVAC systems.

\subsubsection*{Mobility}

Electric vehicles (EV's) and their charging stations provide both new challenges and opportunities. The intermittent load and (combined) large battery provide excellent storage and/or passive balancing opportunities. However, EV arrival and departure times are beyond the control of the charging station, which presents new challenges in charging and discharging batteries at the right time.
In this section, we discuss problems in the control of charging stations, EV battery optimization (sometimes referred to as \textit{powertrain control}), and navigating EV's in a city whenever they need to charge.

In the mobility domain, most interest has been on deep reinforcement learning methods, disregarding tabular approaches. 
Primary objectives are to meet the energy demands of electric vehicles within a desired time, while maximizing the profits of a charging station \cite{athanasiosOpti2022, feifeicuiMult2023, jieliuPric2022, wangModi2022, xianhaoshOnli2022, liCons2020, wangRein2021}. 
This is usually done in a single-agent fashion, where a single charging station is controlled or a single agent controls multiple charging stations. However, some works study cooperative multi-agent settings where more charging stations are considered \cite{jieliuPric2022, xianhaoshOnli2022}.
In the mobility domain, action spaces typically consist of the power output of the charging stations, ranging from a simple total power output \cite{feifeicuiMult2023, athanasiosOpti2022}, to a more detailed per-car output \cite{jieliuPric2022, xianhaoshOnli2022}. 
More elaborate methods focus on energy price prediction \cite{xianhaoshOnli2022, wanMode2019, liCons2020}, decide on the energy selling price \cite{jieliuPric2022, wangRein2021}, include power-generating systems such as solar panels and wind \cite{feifeicuiMult2023}, or explicitly consider the load on the macro-grid \cite{feifeicuiMult2023}.
It has also been suggested that systems that include varying energy prices implicitly optimize to balance the grid \cite{wangRein2021}.
Although most of the works focus on commercial charging stations, one can also optimize the charging station at home, using the car as a battery and selling the electricity back to the grid \cite{wanMode2019}. 

In addition to controlling charging stations, other works focus on navigating electric vehicles more efficiently to a charging station. 
Here, an aggregator system observes city traffic, charging station data, and electric vehicle data and dispatches the most optimal routes to vehicles that need charging \cite{leeDeep2020, q.xingGrap2023}. Alternatively, the reinforcement learning agent may be part of the vehicle itself, navigating it to the most appropriate charging station \cite{qianDeep2020}. 
Various works have also employed reinforcement learning for optimization of EV or hybrid car battery use \cite{biswasReal2019, duDeep2020, liDeep2024a, yaoHybr2022, zhangEner2023}. Furthermore, reinforcement learning has been shown to be capable of helping design new electric vehicles \cite{matallahRein2023}.

Finally, while most research interest is focused on electric vehicles, some works consider hydrogen-fueled cars as sustainable modes of transportation. 
Here, reinforcement learning can help optimize a refueling station with on-site production. The production site is tasked with optimizing revenue by selling hydrogen and balancing the grid, while keeping up with the refueling demands \cite{jiangOpti2024}.

\subsubsection*{Industry}

Industrial consumption often has a distinct consumption profile and often a high volume of energy consumption. 
For instance, data centers are, like houses \& offices, primarily concerned with keeping a building within an acceptable temperature while minimizing long-term energy cost. Here, cooling is again considered crucial to control as close to 50\% of the total energy used in data centers is used for cooling \cite{liTran2020}.
Reinforcement learning has been applied to cooling installations in data centers \cite{biemannData2023, liTran2020}, with Deepmind installing model-based reinforcement learning systems in some of their centers \cite{lazicData2018}.
Reinforcement learning has also been applied in data centers for job scheduling in order to save energy \cite{yiEffi2019, farahnakiaEner2014}, or to create a large system controlling both the scheduling and cooling systems \cite{chiJoin2020, ranOpti2023}.

In greenhouses, various studies optimize the amount of supplementary lighting \cite{afzaliOpti2021} or humidity, cooling and CO$_2$ levels \cite{ajagekarEner2023, zhangRobu2021, morcegoRein2023}.
Other works focus on managing industrial cooling installations \cite{weigoldMeth2021} or cold storage facilities \cite{park502023}.

Reinforcement learning is applied in manufacturing facilities to efficiently power down and restart idle machines, to conserve energy without interfering with production capacity \cite{huangDema2019, loffredoRein2023b, loffredoRein2023}. 
We note that reinforcement learning may find its way into many industrial control operations \cite{elavarasanCrop2020, kazemeiniIden2023, liuRein2022, perez-ponsDeep2021, serranoDeep2019, zhouSmar2022}, but these are not necessarily related to (sustainable) energy and as such are outside the scope of this survey.

\vspace{0.2cm}

The complexity of sustainable energy problems in industry extends beyond mere scheduling and control operations. For certain industries, particularly in the chemical sector, sustainable energy consumption requires not only optimizing energy usage but also sourcing inputs sustainably. Take, for example, the Haber-Bosch process, which relies on hydrogen for ammonia production, typically derived from steam reforming of fossil methane. As discussed in the previous section, reinforcement learning holds promise in sourcing hydrogen sustainably, which could serve not only storage purposes but also as an input chemical for the chemical industry \cite{abiolaNove2023, purnamiDoub2024, liData2021, liCoor2021, sangHydr, tangRein2024}. Furthermore, reinforcement learning has been used to explore new synthesis pathways \cite{kochReinforcementLearningBioretrosynthesis2020}, potentially enabling alternative, sustainable routes. Additionally, reinforcement learning can improve the design of (chemical) catalysts to improve both energy efficiency and input utilization \cite{steinmannAutonomousHighthroughputComputations2022, yoonDeepReinforcementLearning2021, lan2021discovering}.

\begin{figure*}[ht]
    \centering
    \includegraphics[width=1\textwidth]{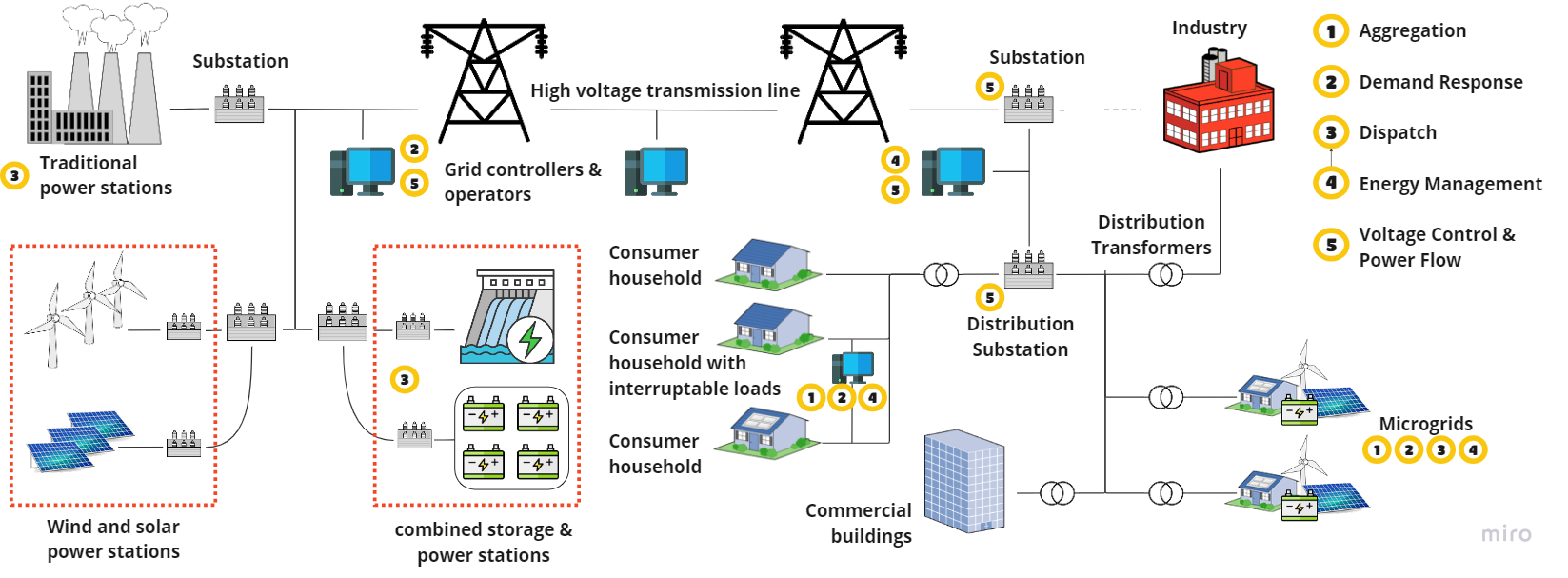}
    \caption{Overview of electrical grid connectivity and control possibilities. {\it Power stations} (left), e.g. traditional plants, distributed energy resources, battery farms, hydrodams, etc., feed electricity into the network. {\it High-voltage transmission lines} (top) then route energy over long distances, facilitating transmission between generation sources and distribution substations. Along the way, the grid contains {\it substations} to ensure the safe routing and operation of electricity, serving as junction points where voltage levels are adjusted and power flow is managed. {\it Distribution substations} further reduce the voltage for distribution to end-users (right), such as households, commercial buildings, and industrial facilities. At these substations, electricity is directed to distribution transformers, which further adjust voltage levels for local distribution. {\it Microgrids} (bottom right) and controllers provide localized energy management solutions, effectively separating local control from the main grid control problem.}
    \label{fig:Overview_electrical_grid}
\end{figure*}

\subsection{Electrical Grids} \label{sect:grids}

We will now turn to the use of reinforcement learning in energy {\it transmission}, focusing on electrical grids. This is the sustainable energy area where reinforcement learning has seen most applications, both in main grids and microgrids. The central challenge is to match supply and demand/load, while retaining grid stability and optimizing energy/cost efficiency. This challenge becomes especially pronounced with sustainable energy sources, since their production profiles are generally uncertain and highly variable (known as {\it intermittency}) and their locations are spread out ({\it distributed}).

The overall structure of the electricity grid is depicted in Fig. \ref{fig:Overview_electrical_grid}. On the left of the figure, we see large-scale energy production and storage facilities that can release energy into the main grid. This includes renewable sources, such as wind and solar energy, and storage stations, such as hydrodams and batteries, as well as -- for now -- traditional (fossil) power stations. These sources enter energy into the main grid, where it is transported over longer distances through high-voltage transmission lines (top part of the figure). Energy thereby ends up at the consumer side (right side of the figure), which includes industry, commercial buildings, and residential houses. Here, we also see the appearance of microgrids, which group a subset of the grid into an independently controllable unit.

In this section we discuss the main challenges that appear within the electricity grid, as marked by the yellow circles in Fig. \ref{fig:Overview_electrical_grid}. 

First, we will discuss key economic aspects of electricity grids, which are vital for its operation. This includes 1) the {\it aggregation} of energy production and consumption sites into stable market entities, 2) {\it demand response}, where we attempt to influence consumers to move energy use to off-peak hours, thereby spreading out demand. Afterwards, the grid needs to be actually operated and stabilized, which involves 4) {\it dispatch}, i.e., the timing of actually release of energy into the grid, as well as 5) {\it energy management}, where we utilize storage capacity throughout (mini-)grids to (locally) match supply and demand. Finally, we also need to ensure the 6) {\it stability} of the grid. This involves {\it voltage} and {\it frequency} control, to keep both within prescribed bounds, as well as {\it power flow}, to effectively route energy through the grid. The location in the grid where each of the above challenges primarily appears is visualised in Fig. \ref{fig:Overview_electrical_grid}.

\subsubsection*{Aggregation} 

{\it Aggregation} deals with combining energy production or consumption sources into stable entities that can participate in the energy market ~\cite{Berntzen2022}. Sustainable energy sources typically 1) vary in production profile (`intermittant') and 2) vary in location (`distributed'). To enter these products into the market, we may groups production and storage sites into {\it Virtual Power Plants} (VPPs) \cite{Liu2023d}. A VPP may, for example, combine solar and wind energy with energy from a hydrogen storage plant. From the outside (to the market) this combination gives the appearance of a stable classical power plant, while it internally combines several variable energy resources. Similar principles can be applied on the demand/load side of the grid, where consumers get grouped together into a single market entity. As such, aggregation allows the production, storage and consumption part of the sustainable energy chain (Fig. \ref{fig:taxonomy_overview}, discussed in the previous sections) to enter the transmission grid as economic entities.  

MDP formulations of aggregation tend to define states based on predicted production profiles of sustainable energy sources, the current status of energy storage sites, and the predicted energy demand. All of these are inherently uncertain and are therefore often separately forecasted \cite{lin2020, ji2022, stanojev2023, wang2022b, algabalawy2021, oh2022}. The reinforcement learning agent then needs to decide on how production sources or consumption loads should be aggregated. Reward functions typically model economic viability, which takes the revenue obtained from the market \cite{stanojev2023,wang2022b} and subtracts internal costs, for example due to over- or undercharging of storage capacity \cite{ji2022}, due to buying additional energy at the market when internal resources are insufficient \cite{ji2022}, or due to the activation of a safety shield to keep the system within operational constraints \cite{stanojev2023}. Note that reward functions may also optimize cost at the customer/demand side \cite{oh2022}.

Due to the many factors involved, aggregation problems typically have a high-dimensional state space, which warrants the use of deep reinforcement learning. Both value-based and actor-critic methods have been applied in this context \cite{lin2020, ji2022, chen2021, stanojev2023, wang2022b, algabalawy2021, oh2022}. Aggregation is typically formulated as a scheduling problem of discharging generators \cite{lin2020}, charging/discharging of storage \cite{ji2022, algabalawy2021, oh2022, wang2022b} or activation/deactivation of flexible loads on the demand side \cite{lin2020, ji2022, chen2021}. Some methods adopt a multi-agent approach, where they treat VPPs as a decentralized system \cite{Rezazadeh2022, Orfanoudakis2023}. Other work focuses on the economic participation of aggregators in the market. Examples include finding optimal bidding strategies \cite{stanojev2023, Xu2021, algabalawy2021} or adjusting pricing schemes \cite{Xu2021, oh2022}.

\subsubsection*{Demand Response} 

Although we typically assume that supply has to match a given customer demand, we may also flip the problem and try to influence demand, better known as {\it demand response}.
While Section \ref{sect:consumption} discussed demand response from the point of view of consumers, actively deciding on local behavior, this section discusses the demand response programs that would entice this behavior, typically through pricing mechanisms \cite{Vardakas2015}.
The overall objective is to minimize energy costs \cite{Fraija2024, Avila2022} or reduce peak demand, which enhances grid stability by spreading out energy consumption over time \cite{bahrami2021, wang2020, oh2022}. 

Reinforcement learning solutions for demand response programs usually model states based on energy demand, grid load profiles, and (renewable) energy availability ~\cite{bahrami2021, wang2020, oh2022, chen2021}. The chosen actions then dynamically adjust pricing schemes ~\cite{Fraija2024, Avila2022}, while the reward function may trade-off the benefit of peak demand reduction with the cost of interrupting end-consumer loads~\cite{bahrami2021, wang2020, chen2021}. Demand response schemes are often implemented in aggregated forms \cite{bahrami2021, Fraija2024} (see previous section), which can entail a multi-agent setting \cite{bahrami2021}. Both value-based \cite{wang2020, chen2021, Avila2022} and policy-based \cite{Fraija2024} methods have been applied.

\begin{figure}[t]
    \centering
    \includegraphics[width=\linewidth]{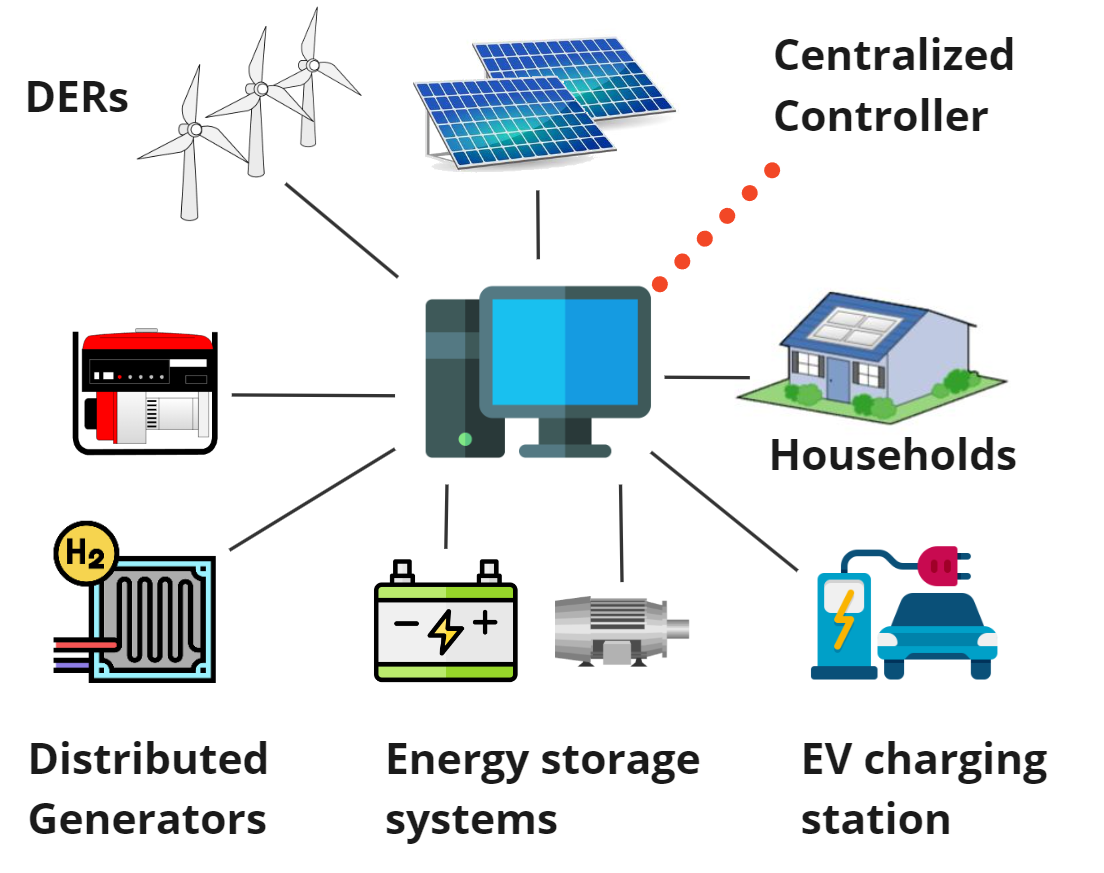}
    \caption{Schematic illustration of a {\it microgrid}: localized energy management within a specific geographic area or facility (e.g., a commercial/industrial building). Microgrids integrate different energy production sites, storage sites, consumer/prosumer households, electricle vehicle charging stations, etc. Energy management within the minigrid is optimized through a central microgrid controller. DER = Distributed Energy Resources.}
    \label{fig:Overview_microgrid}
\end{figure}

\subsubsection*{Dispatch} 

{\it Dispatch} refers to the actual release of energy resources into the electricity grid. This may involve release from sustainable production sites (e.g., a solar farm), storage sites (e.g., a hydrodam) as well as from traditional fossil plants (which are still indispensable in the current energy landscape). Dispatch has always been a challenging problem, since energy resources are spread out, demand profiles are uncertain, and storage capacity is constrained \cite{Li2023, Uddin2023, Dalal2016}. However, the dispatch challenge gets aggravated with the increase in sustainable energy sources, since their production profiles are much more uncertain and variable.

MDP formulations of dispatch problems typically include (partially-observable) state variables from multiple locations over the grid. These may, for example, include current energy generation levels, demand forecasts, state-of-charge of storage systems, and other grid operating conditions~\cite{Li2022b, liu2023c, Lei2020, Sage2023, Dalal2016}. The available actions may then adjust the output of controllable energy production resources~\cite{Li2022b, liu2023c, Lei2020, Sage2023, Dalal2016} or energy storage sites~\cite{Li2022b, liu2023c}, as well as manage connections within the grid ~\cite{Dalal2016}. Finally, objective/reward functions try to match the energy demand while optimizing for economic cost, energy efficiency and/or reliability \cite{Marzbani2024}. Importantly, while classic {\it economic dispatch} \cite{Nghitevelekwa2018, Xia2010} focuses on energy efficiency and cost, {\it generation dispatch} (also known as {\it renewable integration dispatch}) explicitly aims to minimize dispatch from undesirable sources (such as fossil fuels)~\cite{Guo2022}.

Several papers have applied deep reinforcement learning in dispatch optimization. Approaches include both value-based~\cite{Sage2023} and policy-based methods~\cite{Sage2023}, with actor-critic methods (which combine the two) being the most popular~\cite{Li2022b, liu2023c, Lei2020}. The dispatch problem may also be formulated as an {\it hierarchical} MDP, where the effect of certain actions extends over a longer timescale, allowing the agent to better balance short- and long-term energy demands~\cite{Dalal2016}.

\subsubsection*{Energy Management}

{\it Energy management} involves the effective operation and stabilization of the grid itself. For example, in a microgrid we might temporarily have excess energy which we can either store in a battery for later use \cite{Li2022b, ji2022} or sell back to the main grid ~\cite{Monfaredi2023, Harrold2021}. Energy management becomes extra challenging with sustainable energy sources, since their production is highly variable and distributed (e.g., many houses in a minigrid might have solar panels). Note that energy management is a pervasive and broad term, and it is often addressed concurrently with other problems from this section. 

Energy management MDP formulations typically model the state based on the availability of energy resources, energy demand, energy prices, and storage capacity
\cite{Rezazadeh2022, Charbonnier2023, Cui2023, Monfaredi2023, Caputo2023, Liu2023a, Zhou2021b, DBarbero2020}. The action space may consist of load scheduling (i.e., when will a certain demand get activated)  \cite{Charbonnier2023}, storage operations (i.e., charging or discharging a battery) \cite{Charbonnier2023, Harrold2021, Rezazadeh2022, DBarbero2020}, and grid interactions (e.g., trading surplus energy to the main grid \cite{Rezazadeh2022, Harrold2021, Monfaredi2023} or changing the output of a generator \cite{Cui2023, Liu2023a, Zhou2021b, Dalal2016, DBarbero2020}). Reward functions typically need to balance multiple objectives, such as minimizing costs \cite{Rezazadeh2022, Charbonnier2023, Cui2023, Monfaredi2023, Caputo2023} and maximizing the use of renewable energy \cite{Rezazadeh2022, Charbonnier2023}. In addition, it might also consider objectives that reduce battery charge/discharge operations \cite{Charbonnier2023, Monfaredi2023}, ensure grid stability \cite{Liu2023a, Zhou2021b}, or consider an environmental cost \cite{Charbonnier2023, Cui2023, Caputo2023}.

To effectively address these challenges, the majority of approaches implement a deep reinforcement learning framework~\cite{Cui2023, Caputo2023, DBarbero2020, Monfaredi2023,Rezazadeh2022}. In all cases, considerable attention is given to the detailed modeling of the grid configuration to enable realistic simulations~\cite{Rezazadeh2022, Charbonnier2023, Cui2023, Monfaredi2023, Caputo2023, Liu2023a, Zhou2021b, DBarbero2020}. Some methods also utilize the underlying graph structure of grids in their solution approach~\cite{Caputo2023, Zhou2021b}. Although most methods approach the problem from a centralized perspective~\cite{Cui2023, Liu2023a, DBarbero2020}, several studies have also shifted to a multi-agent formulation. For example, one may include individual prosumers in a microgrid as separate agents~\cite{nweye2023}, integrate energy suppliers as decision-makers~\cite{Ghasemi2023, Zhou2021a}, or consider multiple microgrids as agents connected to the same distribution line~\cite{Monfaredi2023, Charbonnier2023}.

\subsubsection*{Voltage/Frequency Control \& Power Flow}

During grid operation, we also need to ensure that both the power line {\it voltage} and the power line {\it frequency} stay within safe bounds. This problem becomes more challenging with sustainable energy sources, especially due to their uncertain output profiles~\cite{Sun2019}. Voltage and frequency may be addressed by primary controllers (which make real-time adjustments at the substation level) or secondary controllers (which make longer timescale corrections across larger sections of the power grid) \cite{zhou_vc}.

MDP formulations of primary power/frequency control define states based on physical variables such as voltage magnitudes, phase angles, power flows, and system frequency \cite{Stanojev2020, Zhang2021}. For primary power/frequency controllers, the action space includes adjustment of reactive power output, switching of capacitor banks, or modification of tap positions of transformers. The action space of the secondary controllers then typically adjusts the setpoints of these primary controllers \cite{Diao2019, Duan2020}. Finally, reward functions aim to maintain voltage/frequency levels within a predefined safe range, while ensuring power quality (i.e., keeping fluctuations low) \cite{Stanojev2020, Diao2019, Duan2020, Zhang2021}. Note that additional objectives may also be included, such as minimization of the energy loss on transmission lines \cite{Diao2020}. 

Deep reinforcement learning approaches for voltage control have often taken an actor-critic approach, where the actor adaptively sets the continuous action space voltage setpoints \cite{Stanojev2020, Duan2020, Zhang2021, Diao2020}. However, value-based approaches have also been tried and generally match or outperform rule-based baselines \cite{Diao2019, Duan2020, Zhang2021}. Larger networks of primary controllers can also be modeled as a decentralized multi-agent reinforcement learning problem, thus providing an alternative to secondary controllers~\cite{Xu2023}. Overall, reinforcement learning methods for voltage and frequency control have shown promising results in simulation, but their real-world value, of course, depends on the quality of the underlying simulation models.

Finally, {\it power flow} optimization \cite{Risi2022} is a bridging field between high-level energy management (previous section) and low-level voltage/frequency control (this section). Power flow involves optimization of a complete electrical system, which may involve voltage control but also transmission line switching \cite{Zhou2021b} or dispatch from energy generators \cite{Liu2023a, Zhou2021b}. Similar to energy management methods, power flow approaches may also utilize the underlying graph structure of the problem \cite{Lehna2023, vanderSar2023, Zhou2021b}. As such, power flow optimization overarches the space between high-level energy management and lower-level voltage/frequency control, where the exact boundaries vary between papers.

\section{Reinforcement Learning Challenges} 
\label{sec:5}

While the previous section summarized the literature from the sustainable energy point of view, we now revert our viewpoint and focus on the overarching reinforcement learning themes that appear throughout the literature. These central reinforcement learning topics include multi-agent RL, partial observability, model-based RL, offline RL, and safe RL. The connection between both directions (sustainability problems and reinforcement learning solution methods) is visualized in Table \ref{tab:sec5}. Clearly, these reinforcement learning topics are not mutually exclusive, e.g. one might deal with a partially-observable multi-agent setting.

\setlength{\extrarowheight}{10pt}
\begin{table*}[ht]
\centering
\caption{An estimate of how popular certain reinforcement learning challenges/topics are within different sustainable energy fields field. This table only indicates to what extent the current literature, included in our survey, deals with varying reinforcement learning problems. As such, this table may indicate a research gap for those interested in sustainable energy reinforcement learning problems.
}
\label{tab:sec5}
\begin{tabular}{l ccccc}
\toprule
 &
 Multi-Agent RL &
 Partial Observability &
 Model-based RL &
 Offline RL &
 Safe RL \\ \midrule
Generation  & \transparent{0.5}{\ding{52}} & \transparent{0.75}{\ding{52}} & \transparent{0.25}{\ding{52}} & \transparent{0.25}{\ding{52}} & \transparent{0.5}{\ding{52}} \\
Storage     & \transparent{0.1}{\ding{52}}  & \transparent{0.1}{\ding{52}} & \transparent{0.1}{\ding{52}} & \transparent{0.1}{\ding{52}} &  \transparent{0.1}{\ding{52}}\\
Consumption & \transparent{0.75}{\ding{52}} & \transparent{1.0}{\ding{52}} & \transparent{0.25}{\ding{52}} & \transparent{0.25}{\ding{52}} & \transparent{0.25}{\ding{52}} \\
Grids       & \transparent{0.25}{\ding{52}} & \transparent{0.25}{\ding{52}} & \transparent{0.1}{\ding{52}} & \transparent{0.1}{\ding{52}} & \transparent{0.3}{\ding{52}} \\
\bottomrule
\end{tabular}
\end{table*}

\subsection*{Multi-agent RL}

Conventionally, reinforcement learning is concerned with finding the optimal policy of just a single agent.
However, many problems are modeled as multi-agent problems where any number of agents act in the same environment, jointly affecting its state space \cite{wongDeep2023a}.
As a single agent is now no longer solely affecting the transition function, each agent will perceive a higher amount of unpredictability, increasing the state space and destabilizing the learning process.  
Sometimes agents may communicate, or cooperate. 
However, in some scenarios, there may be delays in communication, agents may not wish to share preferences, or there are privacy concerns \cite{Charbonnier2023}.
Furthermore, distributed multi-agent environments may also be used as a means to model large state-action spaces. 
Distributed state-action spaces are smaller and easier to train on.
A popular method for large multi-agent state spaces is to train algorithms according to \textit{centralized training, decentralized execution} (CTDE). 
Here, multi-agent systems can exchange certain attributes during training, to increase efficiency, but this interaction is removed in deployment.

In sustainable energy reinforcement learning research, problems are often modeled as multi-agent problems, often in a collaborative manner.
Frequently, these studies resort to improved trainability by splitting up larger state-action spaces over multiple agents, or CTDE.
This is for example done in building control, controlling multiple appliances or multiple zones \cite{fuOpti2022, yuEner2022, yuMult2021, dengOpti2022, bladData2022, luDema2019}, EV charging stations \cite{xianhaoshOnli2022, jieliuPric2022}, hydrogen refueling stations \cite{jiangOpti2024}, energy management in grids \cite{Charbonnier2023, Zhou2021a}, voltage control \cite{Xu2023} and power flow optimization \cite{bahrami2021}, 
as well as in energy generation. Here, for example, windparks may be modeled as distributed multi-agent systems, splitting up the state-action space \cite{bui_distributed_2020, sarkarMult2022}.

In demand response, systems that control multiple buildings \cite{vazquez-caMARL2020a} are also a natural match for multi-agent reinforcement learning. 
\cite{citylearn_challenge_2020, citylearn_challenge_2021, citylearn_challenge_2022}.
While more advanced multi-agent systems are infrequent in the literature, one study models its multi-agent system of cells in a hierarchical structure, dividing the wind farm into segments. Within these segments, agents are able to exchange information with relevant nearby agents \cite{ge_maximizing_2021}.

\subsection*{Partial observability}
A standard Markov Decision Process (Section \ref{sec:3}) assumes perfect information. However, most real-world problems are actually Partially-Observable Markov Decision Processes (POMDPs) \citep{spaan2012partially}. {\it Partial observability} refers to the fact that the current observation often does not capture all information of the ground-truth state of the system \citep{jaakkola1994reinforcement}.
For example, in a first-person view navigation task, the current observation does not provide information about the environment behind us, and in a card game one agent may not know the hidden cards of their opponent.
This partial observability may be mitigated by incorporating additional information from historical observations, i.e., a form of {\it memory}. However, taking our entire history into account quickly becomes computationally infeasible. 

Partial observability has been studied extensively in the reinforcement learning literature, for example for policy estimation \citep{wierstra2010recurrent,parisotto2020stabilizing}, value function estimation \citep{hauskrecht2000value, hausknecht2015deep}, and the dynamics model \citep{chiappa2016recurrent,chen2022transdreamer}. 
Key methods in deep learning that are used to address partial observability include windowing/framestacking \cite{linMemo1992}, recurrent neural networks \citep{medsker1999recurrent, hochreiter1997long, gu2021efficiently}
, transformers \citep{vaswani2017attention}
, external memory methods \citep{graves2014neural} and state space models \cite{guEffi2022, luStru2023}. 
Another approach for addressing POMDPs involves utilizing belief systems, which maintain a probabilistic representation of the agent's current state based on past observations and actions~\cite{Cass1995}.

In sustainable energy research, partial observability is prevalent. Many problems involve time-series data, such as weather data \cite{yuEner2022, zhangRobu2021, chou_maximum_2019, correa-jullian_operation_2020, zhang_two-step_2022, Li2022a}, dynamic electricity prices \cite{fuEDDQ2023, wangRein2021, riemer-sorensen_deep_2020}, and occupancy data in buildings \cite{dengRein2021, parkLigh2019}---all based on processes that are difficult to fully observe. 
  
A number of studies simply ignore the partial observability in these problems \cite{yuEner2022, coraciOnli2021, fuOpti2022, wangModi2022}, treating the problem as a stochastic problem, which is sometimes effective.
Others try to address the problem by engineering historical information for specific features \cite{liCons2020, wanMode2019, liReal2020, xianhaoshOnli2022, zeng_real-time_2022}, or future (predicted) features \cite{luDema2019, gaoComp2023}.
This approach may enhance performance, but is dependent on the ability of the engineer to construct the right set of features. 
Other methods found in the literature use frame stacking \cite{jeenLow2023}, or include LSTM layers in the neural network \cite{liuMult2022b, xianhaoshOnli2022, wanMode2019, yiEffi2019, biemannData2023, sarkarMult2022, Lei2020, Zhou2021a}.

Belief systems are only rarely used in sustainability research to deal with partially observable problems. In one study, it is used exclusively to model beliefs about the states of other agents in multi-agent formulations \cite{Charbonnier2023}. The belief states are used to model the possibility of communication failures between different households that need to cooperate within an electrical grid. 
The limited use of this method aligns with the trend in general reinforcement learning, where belief systems are becoming less common due to their inability to scale and the requirement of inserting a lot of prior knowledge.

\begin{figure*}[ht]
    \centering
    \includegraphics[width=1\textwidth]{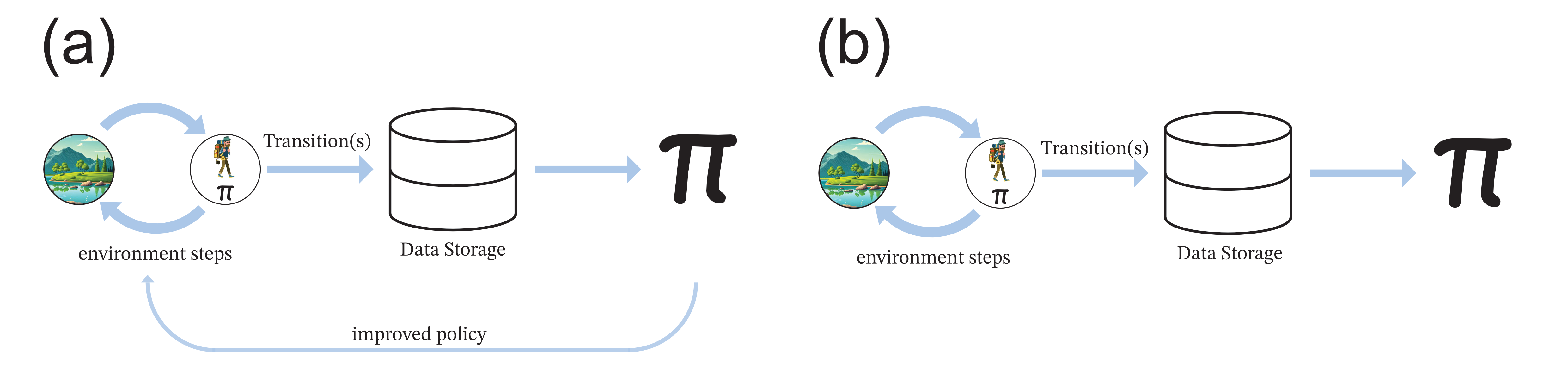}
    \caption{(a) Online Reinforcement Learning: The improved policy is continually updated and integrated into the system, directly influencing the agent's behavior. (b) Offline Reinforcement Learning: Transition data is collected in advance, and a policy is derived from this data without the typical feedback loop seen in Online Reinforcement Learning (Based on \cite{levine2020offline}, figure 1).}
    \label{fig:17}
\end{figure*}

\subsection*{Model-based RL}

While {\it model-free} reinforcement learning methods directly learn a solution (value or policy) from observed transition data, {\it model-based} reinforcement learning \citep{moerlandMode2023} instead first learns the {\it transition and reward dynamics} ($p(s'|s,a)$ and $r(s,a,s')$, see Section \ref{sec:3}) of the problem. This effectively recovers an internal {\it model} in the agent of the true environment MDP. 
The agent can subsequently {\it plan} using its learned model, without consulting the environment, to update its solution \citep{sutton1990integrated}, which can greatly increase sample efficiency \citep{deisenroth2011pilco}. 
Note that this approach has connections to many other challenges in this section: model-based RL is a common approach in the offline setting \citep{kidambi2020morel,yu2020mopo}, planning over a model may help to ensure safety guarantees \citep{berkenkamp2017safe,thomas2021safe}, and the model itself often suffers from partial observability (all discussed in different parts in this section).

Some methods first learn a dynamics model from an offline fixed dataset, and subsequently use the model as a simulator for training \cite{yiEffi2019}.
Other methods build or improve their models during training in a simulator and use these models for planning \cite{faridiAdva2024}. 
In addition to planning, trained models can also be used to generate additional training samples, in addition to the training samples generated by the real world or a simulator \cite{zhangRobu2021}.

In the building control domain, a direct comparison has been made between model-free methods and their model-based counterparts, based on MBPO \cite{janner2019trust}. 
Here, the authors demonstrate that model-based methods generally converge faster and may potentially deliver superior final performance compared to model-free methods \cite{gaoComp2023}.
This importance of convergence speed is underlined by Jeen et al.~\cite{jeenLow2023}, who note that it is infeasible to construct simulators for every building to pre-train on. 
As previously mentioned, model-based reinforcement learning methods are also found in Google data centers, where a model is trained for planning using model predictive control \cite{lazicData2018}.

Note that model-based approaches excel when the transition model is exactly known, such as in Chess or Go \citep{silver2018general}. However, when the transition model is learned from data, model-based reinforcement learning can become unstable, due to model uncertainty \citep{wu2022plan} and accumulating errors during planning \citep{janner2019trust}. However, model-based methods have been reported to achieve state-of-the-art performance \citep{hafner2023mastering}, and could be a valuable approach in various sustainability challenges.

\subsection*{Offline RL} \label{sect:rl-ch-offline}

Another area of reinforcement learning that is used frequently in sustainable energy is offline reinforcement learning. Vanilla {\it online}
reinforcement learning methods assume access to an environment which we can continuously query for new transition and reward data. However, for many real-world problems we do not have a (good) simulator because we do not know the transition function, and we cannot afford continuous (exploratory) interaction with the real system either | for example, because it needs to supply customers and a break in service is unacceptable for business or ethical reasons. 
In those cases, we may be able to obtain a batch of transition data (data in the form of \{state, action, reward, next state\}) from the real system and want to find an improved policy solution from that finite dataset. This is known as {\it offline} reinforcement learning \citep{levine2020offline, agarwal2020optimistic, fujimoto2021minimalist, janner2021offline}. 

Offline reinforcement learning introduces several challenges. In the model-free setting, we primarily need to rely on {\it off-policy} reinforcement learning methods (see Section \ref{sec:3} -- not to be confused with {\it offline}) \citep{kumar2020conservative}) since off-policy methods can find an improved policy given arbitrary transition data. As an alternative, we may also take a model-based approach, where we first learn a model of the MDP, after which we can apply any reinforcement learning algorithm through planning in the model \citep{kidambi2020morel,yu2020mopo}. The main challenge originates from uncertainty due to limited data \citep{Bai0YDG0W22, NEURIPS2021_3d3d286a}: we want to prevent our solution from diverging too much from the observed data region, since our predictions will then become uncertain and our obtained solution may become (very) suboptimal.

In the offline reinforcement learning settings, we find very little literature. 
In the building control domain, Blad \textit{et al.} \cite{bladData2022} have fully trained their reinforcement learning agent from transition data.
Other works attempt to solve the offline challenge with a model-based approach.
Here, the system dynamics (the transition function $p$ and the reward function $r$) are trained in a supervised manner using real data. This model can then be used as an environment upon which an reinforcement learning agent is trained \cite{faridiAdva2024, yiEffi2019}.
The small amount of literature indicates a potential research gap. Especially in energy systems, companies may often have been collecting large amounts of data on their current (non-RL) systems. 
If this data were to be accessible to the research community, we expect offline reinforcement learning to become an important cornerstone in the research field.

In the literature, we often observe problems where a simulator exists, but the state space is augmented by an offline time-series dataset. This is common for frequently used state attributes, such as weather data or electricity prices, to create a more realistic simulation \cite{riemer-sorensen_deep_2020, xu_deep_2020, shresthamali_adaptive_2017, Liu2023b, Cuadrado2023, cao2020, liu2023e, DBarbero2020}. These state attributes are then assumed to remain beyond the influence of the agent.
Although we do not consider these problems \textit{offline} reinforcement learning, where we only deal with a finite dataset of transitions, we want to make the reader aware of this practice. In particular because it is widespread in the sustainable energy field (because weather and energy prices are common), and, similarly to offline reinforcement learning, these problems may be limited in their possible generalization. For instance, given only data of the past year, it does not guarantee us to generalize to any potential energy price of the future. Furthermore, as in offline reinforcement learning, the problems require us to keep a held-out test set to validate performance.

\vspace{0.2cm}

It is worth noting that there may be instances of terminological ambiguity. Some papers~\cite{jia_reinforcement_2021, zhuangData2023} use to the term "offline reinforcement learning" to refer to the offline-state of an offline/online deployment process, rather than offline reinforcement learning in the context discussed in this paragraph.

\subsection*{Safe RL} \label{sect:rl-ch-safety}

{\it Safety} is a key topic in all real-world applications of reinforcement learning \citep{pecka2014safe,garcia2015comprehensive,gu2022review}. Many real-world systems are vulnerable: they need to operate within safety boundaries to avoid malfunction or collisions. When we learn on such a system, we therefore first of all need to ensure that the system will not break or cause harm, i.e., stay within the prescribed bounds. 
Some applications may lend themselves to being layered by different control agents. Whenever the state lands outside of predefined safe bounds, the controls may be shifted one layer up to a safer, but perhaps less efficient, control agent.
This kind of setup would unfortunately not always be possible, and we prefer our most efficient agents to behave safely as well. 
As such, many reinforcement learning methods have been designed to tackle this problem \citep{junges2016safety, alshiekh2018safe, lutjens2019safe, cheng2019end, yang2023safety}. Most of these methods have been studied in the robotics community, since these systems are generally vulnerable \citep{brunke2022safe}.

A common approach to add a first layer of safety involves pre-training a learning algorithm on a virtual representation of the robot to narrow its action space prior to real-world deployment \cite{Zhou2021b, tubeuf_increasing_2023}.
While this approach may accelerate learning post-deployment and mitigate the risk of instabilities, it does not entirely eliminate the possibility of system failures. 

In {\it safe RL}, however, the goal is to completely eliminate the risk of such failures. This could be achieved by constraining the action space through hard constraints~\cite{bui_distributed_2020, stanojev2023, Werner2023}. However, such constraints may lead to suboptimal behavior. Alternatively, the environment may be modified.
Here, hard constraints can be imposed via a cost function \cite{stanojev2023}, allowing the agent to optimize only among policies that fall below a certain maximum cost threshold \cite{zhangRobu2021}.

Another approach to integrate safety concerns is to use a model-based reinforcement learning approach based on a dynamics model \cite{anCLUE2023}. Here, the model incorporates uncertainty estimates that are learned using a Guassian process, providing the agent with not only an expected return but also an uncertainty estimate during the planning phase. 
In a similar fashion, another work \cite{liCons2020} proposed to enhance the robustness of an agent in bad scenarios, or, in other words, reduce the uncertainty of estimates in the face of bad rewards. This is done by modifying the replay buffer during training to retain only transitions with poor rewards while discarding those with high rewards.

\section{Benchmarks, baselines and performance metrics} \label{sec:6}
So far, we have discussed various sustainable energy problems that lend themselves to reinforcement learning solutions (Section \ref{sec:4}), and discussed various reinforcement learning areas that are prevalent in the current literature (Section \ref{sec:5}).
Successful progress in any field of research often benefits from standardization. 
In this section, we examine different benchmarks (environments), baselines (algorithms), and performance metrics used throughout the sustainable energy landscape within the field of reinforcement learning.

\subsection*{Benchmarks}

Standardized benchmarks are a cornerstone of the progress in machine learning, facilitating the comparison of different methods on identical tasks and enabling a fair assessment of approaches. In reinforcement learning a significant advancement in this regard was realized with the introduction of Gym \cite{openai_gym_original} and subsequently Gymnasium \cite{gymnasium}, which standardized an API for reinforcement learning environments. 

The field of sustainable energy encompasses a broad spectrum of challenges, as highlighted in Section \ref{sec:4}.
Despite (or because of) this diversity, numerous efforts have been made to create standardized environments.
For ensuring wide usage and compatibility, ideally, they should also meet specific needs, including broad scope coverage, active maintenance, and integration with common frameworks like Gym/Gymnasium.
An overview of these environments is given in Table \ref{tab:benchmarks}, in which some of the main simulators are given that are ready to use in a Python-based reinforcement learning pipeline.

Notable in Table \ref{tab:benchmarks}, the building control problem has the widest variety of simulators available, possibly due to the popularity of the EnergyPlus building simulator \cite{crawleyEner2001}.
Of further particular interest is SustainGym \cite{yehSust2023}, which occurs in multiple settings. SustainGym is a library consisting of multiple sustainable energy Gymnasium environments that are, at the time of writing, still actively maintained. 
We encourage researchers entering the field to investigate the use of SustainGym for their studies and to consider developing new environments for its framework.
Some of these benchmarks originate in challenges hosted in the past, such as the \textit{CityLearn challenge} \cite{citylearn_challenge_2020, citylearn_challenge_2021, citylearn_challenge_2022} or the \textit{L2RPN (Learning to Run a Power Network) challenge} \cite{Marot2020, Marot2021, Marot2022}.
These challenges serve as open benchmarks for continued submission beyond the original deadlines of these challenges.

In the power transmission domain, the \textit{grid2op} python package \cite{grid2op} (which replaces the now-defunct pypownet \cite{lerousseau2021design} package) offers 
functionalities for simulating and benchmarking power grid scenarios 
and forms the foundation of the L2RPN challenge. 
Other recent environments include \textit{gym-anm} \cite{Henry2021a, Henry2021b}, which addresses similar power grid management tasks, but tailored for \textbf{A}ctive \textbf{N}etwork \textbf{M}anagement in distribution networks; or \textit{PowerGridWorld} \cite{Biagioni2022} which is specifically tailored for multi-agent implementations. All of these are compatible with Gym or Gymnasium, and are openly accessible through GitHub. 

\begin{table}[ht] \label{tab:benchmarks}
\caption{Overview of environments/simulators in the realm of sustainable energy. These simulators are open source available on GitHub and often available via a Python API.}
\centering
\label{tab:env_table}
\begin{tabular}{llp{3cm}}
\toprule
\multicolumn{2}{c}{\it Sustainable Energy Area}         & \it Simulator \\ \midrule
\textbf{Generation}  

& Wind            &   \cite{OpenFast, wfsim}        \\
                                   & Solar           &     \cite{gym-PVDER}      \\

\textbf{Consumption} 
& Buildings       &     \cite{yehSust2023, blumBuil2021, blumBuil2021, findeisBeob2022, nweyeCity2023a, vazquez-caCity2019, zhangCOBS2020, jimenez-raSine2021, scharnhorsEner2021, wangAlph2021a}      \\
                                   & Industry        &    \cite{yehSust2023, tesslerRein2022, samsonovRein2022}       \\
                                   & Electrical vehicle              &      \cite{yehSust2023, yanMobi2022, karatzinisChar2022, orfanoudakEV2G2024}     \\

\textbf{Grids} 
& Grid management &     \cite{Henry2021a, grid2op, lerousseau2021design, Biagioni2022}      \\

\bottomrule
\end{tabular}
\end{table}

\subsection*{Baselines and Performance Metrics}

In reinforcement learning, \textit{benchmarks} define the environment and are used to measure the performance of an algorithm to solve a task. However, for a fair comparison, the performance of other solutions for the same environment is required, which we refer to as the \textit{baselines}. 
Baselines often come in two different flavors: either as alternative non-reinforcement learning-based algorithms or as state-of-the-art reinforcement learning algorithms. In the realm of sustainable energy, the former baselines are usually chosen as classical optimization techniques, which are used in real-world applications. 
In contrast, the latter are usually used to show an improvement over previous solutions or to explicitly compare different approaches on the same environment.

As the field of reinforcement learning in sustainable energy is relatively new, the focus of most presented research lies on developing proof-of-concept reinforcement learning solutions for given problems and comparing them to classic (non-learning) control solutions. These include rule-based controllers, e.g. PID controllers \cite{enyekwe_speed_2023, gaoComp2023, qiuMult2023, jiangBuil2021, jeenLow2023, liTran2020, biemannData2023} or other heuristic controllers \cite{farahnakiaEner2014, yiEffi2019, huangDema2019, loffredoRein2023b}. We also see comparisons with more sophisticated methods such as learned decision trees \cite{xu_deep_2021}, dynamic programming \cite{yang_reinforcement_2021, mitjana2022managing, xu_deep_2020, xu_deep_2021}, and model predictive control \cite{anderliniReal2020, wanMode2019, liCons2020, morcegoRein2023, ajagekarEner2023, jeenLow2023}.
In some cases, an optimal solution is accessible during training. The authors may then compare against an optimal \textit{oracle} baseline that assumes future knowledge to be known (operating in hindsight) \cite{anderliniCont2016, wei_adaptive_2016, jeenLow2023, afzaliOpti2021, qianDeep2020}. In other instances, optimal solutions might be obtained through established methods such as linear \cite{Monfaredi2023, Silva2021} or nonlinear programming \cite{liu2023c}, but are not necessarily capable of dealing with uncertainties. 

In some of the research on sustainable energy, existing reinforcement learning algorithms are used as a baseline. In these cases, often a specialized problem is addressed that benefits from a well-engineered reinforcement learning method \cite{jeenLow2023, xie_wind_2022}.
Reinforcement learning algorithms generally require a high level of engineering and hyperparameter tuning, requiring careful consideration when using alternative algorithms as a baseline. 
As such, it is generally advised to use well-engineered standardized baselines such as {\it Stable-Baselines} \cite{raffinStab2021}, {\it CleanRL} \cite{huangClea2022} or {\it RLlib} \cite{liangRLli2018a}.

\vspace{0.2cm}

In reinforcement learning, the value function, or the expected episodic return, is the predominant metric for evaluating the performance of a learning algorithm, as rewards are designed to resemble some measure of optimality. However, for numerous applications, a single metric alone does not provide a complete picture. Multi-objective reinforcement learning studies this problem \cite{hayes2022practical}. Reliability 
emerges as a significant factor in many energy domains and is therefore sometimes adopted as an additional performance metric \cite{xu_deep_2020, xu_deep_2021}. 
Moreover, classical control methods such as model predictive control suffer from considerably longer inference times compared to reinforcement learning, leading to the emergence of another sensible performance metric not captured by the episodic return \cite{zhang_reinforcement_2020, zeng_real-time_2022}.
Some papers adopt a more sustainability-focused performance metric. This typically includes some form of emissions measure, e.g. the amount of $CO_2$ emitted \cite{Caputo2023} or renewable utilization \cite{Liu2023a}, if it can be calculated within the described system. 
To summarize, multiple performance metrics can be taken into account in the sustainable energy domain. As such, it is important that researchers in the field carefully consider what they want to evaluate.

\section{Discussion and Future Work} \label{sec:7}

Matching supply with changes in demand is one of the major challenges in sustainable energy. New elements have arrived in the energy chain, such as batteries, smart grids, and smart appliances. This has led to a significant increase in the need for control and optimization of -- often interconnected -- decision problems, a topic for which reinforcement learning methods are very well suited. 

The combination of reinforcement learning in sustainable energy is still young, and the richness and fast growth of the landscape has resulted in a scattered field in terms of environments and benchmarks.
A large amount of research in the field uses undisclosed, problem-specific environments or its own hand-crafted environment, often not open-sourcing the source code. 
This leads to redundant efforts and impedes the reproducibility of results.

As such, we believe that all areas in the field would greatly benefit from well-designed and general environments that receive long-term maintenance and can be continuously built up on. A promising software package in this regard is SustainGym \cite{yehSust2023}, which attempts to standardize a variety of different sustainable energy environments.
Such a standardization would allow for easier use of standardized algorithm implementations (such as Stable-Baselines and CleanRL). 
Furthermore, the entry barrier for reinforcement learning engineers would be reduced as they no longer need to focus on building realistic and relevant environments.

We note a flourishing amount of research in the field, often aimed at an initial demonstration of the potential of reinforcement learning. 
Often, well-known algorithms are used, such as tabular methods or DQN. A deeper look into the problem situation and the reinforcement learning literature may well be worthwhile to achieve better results. 
As the field matures, more interdisciplinary teams will arise, knowledge of the energy and algorithms field will integrate, and we expect more breakthrough results to appear.

Specifically, we will discuss some important reinforcement learning methods that may well be important for further progress. Firstly, some papers have accurately identified that the construction of simulators is not always feasible \cite{jeenLow2023}; this may especially be true in the consumption domain, due to the diversity of consumption patterns. 
Model-based reinforcement learning methods may aid sustainable energy challenges as they first build a transition and reward model before the policy is optimized. 
So far, we see relatively little use of model-based methods in the field, suggesting it is an underexplored approach.

Next, when creating or learning a simulator is not feasible, offline reinforcement learning may be an option. 
Over time, large amounts of has likely been collected on various existing operating processes.
This in turn opens the door for offline reinforcement learning, were this data to become available. Offline reinforcement learning is a field of research rich in literature \cite{levine2020offline, janner2021offline} and the developed methods appear to not be explored in the sustainable energy domain.

Generative deep learning \cite{salakhutdiLear2015} has received much attention in the last decade. As stronger methods become available, generative deep learning is likely to aid in modeling system dynamics. This would strengthen the fields of both model-based- and offline reinforcement learning. A promising avenue to investigate currently would be diffusion models~\cite{zhu2023diffusion}.

Safe RL is another research area that is highly relevant for sustainable energy.
In order for applications to get past the proof-of-concept phase, we often require incorporation of safety aspects into the system. 
This is because \textit{a)} we cannot allow disruptions due to exploratory action during the training face in most energy systems, and \textit{b)} even if we were able to perform the training phase outside the live system, all approximate machine learning methods (including deep reinforcement learning) provide some level of generalization. While this is in part a desired trait, this also results in our systems always predicting some output, without us being able to verify whether that output is correct everywhere. 
Most current reinforcement learning research is therefore studied in simulation or non-safety critical applications, but the requirements for real-world deployment are much stronger (e.g. self-driving cars). A main challenge of RL and ML in general is deployment in real-world, safety-critical scenarios, and the energy world will be a prominent example of this.

The research literature reports few cases where reinforcement learning methods are applied in practice. 
This might be due to three reasons. First, the field is generally quite young and there might not have been enough time for (safe) deployment. In addition, the safety concerns themselves might also be a reason machine learning methods have not found their path to deployment yet. 
Finally, deployment by actors in various energy markets may simply not publish their findings in peer-reviewed research, making it harder to accurately assess the status of deployment. 

Another promising area of research is graph-based reinforcement learning.
Graph neural networks \cite{wuGrap2022} in general have received much attention. Yet, in the surveyed literature, it has not been applied to a great extent in the energy domain as part of a reinforcement learning agent. 
However, graph structures appear naturally throughout the sustainable energy landscape. Prime examples would be (electrical) grids and chemical molecules.
Those with interest in graph neural networks and reinforcement learning are strongly encouraged to apply their knowledge in the sustainable energy domain. 

Furthermore, {\it multi-objective} reinforcement learning \cite{hayes2022practical} might also become relevant, since possible reward objectives, such as profit and emission reduction, can be conflicting. Multi-objective RL methods that can adaptively trade-off objectives thereby become relevant. In addition, also {\it multi-scale} approaches hold promise. Especially the electricity grid has many subproblems that appear at different scales, and they all interact for overall grid efficiency. Some research does look at different scales of a problem simultaneously \cite{Cuadrado2023, Xu2021}, and integrating solutions is a promising avenue for future research.

Finally, we point out that much research in the sustainable energy landscape relies on weather and energy price data. Future predictions of this data would be especially relevant for a wide variety of use cases.
While some works attempt to include such prediction mechanics into their models, we believe the sustainable energy field would benefit most from addressing its own problems and using external models, developed in other fields, to address the challenges of time-series prediction \cite{lam2023learning}.
However, systems that accurately take history into account are a promising pathway for reinforcement learning in general. Due to the amount of time-series data in the energy domain, this is especially relevant for sustainable energy research. As such, we believe that the recent advances in state-space-models \cite{guEffi2022, luStru2023} may aid the performance of reinforcement learning agents in a wide variety of tasks discussed in this survey.

\section{Conclusion}

This survey provides a comprehensive overview of the available reinforcement learning approaches to sustainable energy challenges. We first of all observe that the research field has grown rapidly in recent years, and there are many sustainability challenges to which reinforcement learning is applicable. However, we also observe that most papers originate from energy researchers that start to apply reinforcement learning methodology, while reinforcement learning researchers are less present -- probably because they struggle to understand the relevant underlying problems and the way to model them. Therefore, to mature the field, we likely need more interaction between researchers from both communities. A key direction for integration would be the development of better benchmarks, that is, standardized test environments, on which we can test and compare reinforcement learning approaches. Even then, real-world deployment will probably also require the development of new core methodology, most notably in safe and offline RL. In short, this survey identifies a large potential for reinforcement learning to contribute to the sustainable energy transition. Given the urgency in solving these problems, we hope to see the field mature and interdisciplinary work thrive.

\section*{Acknowledgements}
This work was supported by Shell Information Technology International Limited and the Netherlands Enterprise Agency under the grant PPS23-3-03529461.

\bibliographystyle{ieeetr}

\bibliography{references}

\end{document}